\expandafter\def\csname ver@breakurl.sty\endcsname{2013/04/10 v1.40 (skipped for XeTeX)}
\documentclass[hidelinks,onefignum,onetabnum,nohypdvips,juq]{siamonline250211}

\usepackage{amsfonts}
\usepackage{amssymb}
\usepackage{graphicx}
\usepackage{booktabs}
\usepackage{algorithmic}
\graphicspath{{figures/}}

\newsiamremark{remark}{Remark}
\newsiamremark{hypothesis}{Hypothesis}
\crefname{hypothesis}{Hypothesis}{Hypotheses}

\headers{Working precision in GP maximum likelihood}{T. A. Scott}

\title{The Danger of Double Precision: Working Precision as a Hidden Constraint in Gaussian Process Maximum Likelihood%
\thanks{Submitted to the editors 11/19/2026.
}}

\author{Thomas A. Scott\thanks{Department of Aerospace Engineering and Engineering
  Mechanics, The University of Texas at Austin, Austin, TX
  (\email{thomas.scott@utexas.edu}).}
  }

\ifpdf
\hypersetup{
  pdftitle={The Danger of Double Precision: Working Precision as a Hidden Constraint in Gaussian Process Maximum Likelihood},
  pdfauthor={Thomas A. Scott}
}
\fi

\begin{document}

\maketitle

\begin{abstract}
Gaussian process (GP) hyperparameters are routinely fitted by maximizing the
marginal likelihood in IEEE double precision, although for smooth kernels
and noiseless data the covariance matrix is often more ill conditioned than
double precision can resolve. For a squared-exponential kernel on a design
with a Bayesian-optimization cluster, refits over a range of significand
widths, judged against a high-precision reference, show that the
double-precision fit is a different model and that fits converge only once
the width exceeds the base-2 logarithm of the covariance condition number,
about 70 bits. Cholesky failure acts as an implicit conditioning
constraint, caused mostly by rounding of the kernel entries; the
constrained solution predicted from the working precision alone is within
20\% of the best Powell fits at five of the six precisions from 40 to 64
bits, with a local
expansion proved for the idealized problem. Across 43 kernels, designs and
problem settings, including negative controls that double precision already
resolves, the bits needed follow the log condition number at the
reference optimum plus a tolerance-dependent offset, and at the loosest
tolerance double precision
suffices for the Matérn 3/2 kernel but not for the squared-exponential
kernel on clustered designs. Double-precision hyperparameters overstate the
posterior standard deviation near the design cluster and, on a quadratic
objective whose likelihood heads to the flat limit, stall closed-loop active
learning. Promoting the whole likelihood to 80 significand bits recovers the reference accuracy
on every fixed design tested, whereas a fixed jitter trades failures for a
posterior bias.
\end{abstract}



\section{Introduction}\label{sec:intro}

Gaussian process (GP) regression is the standard emulator for deterministic
computer models and the default surrogate in Bayesian optimization (BO)
\cite{Rasmussen2006GPML,Sacks1989DesignAnalysisComputerExperiments,Jones1998EfficientGlobalOptimization,Shahriari2016HumanOutOfLoop}.
Its hyperparameters are usually set by maximizing the marginal likelihood,
which requires a Cholesky factorization of the covariance matrix
$A_\theta = K_\theta + \sigma_n^2 I$ (\cref{sec:background}) at every
evaluation; for smooth kernels and nearly noiseless data $A_\theta$ is
notoriously ill conditioned, and practitioners add a small ``jitter'' or
nugget to the diagonal
\cite{Ranjan2011ComputationallyStable,Gramacy2012CasesNugget,Peng2014ChoiceNugget,Andrianakis2012EffectNugget}.
We ask a question that is usually left implicit:
\emph{is IEEE double precision converged for the maximum-likelihood
hyperparameters themselves?}

That double precision can corrupt the GP likelihood is known:
McCourt and Fasshauer~\cite{McCourt2017StableLikelihoodGRF} removed an
artifact minimum of the likelihood with a stable basis,
Basak et al.~\cite{Basak2022NumericalIssuesMLE} modeled the roundoff noise
in the double-precision negative log-likelihood (NLL) as roughly $\kappa u$
($\kappa$ the covariance condition number, $u$ the unit roundoff) on the
Branin function also used here, and Karvonen and
Oates~\cite{KarvonenOates2023MLEIllPosed} remark that the likelihood can
in practice only be maximized where it ``can be reliably computed''.
Zimmermann~\cite{Zimmermann2010AsymptoticLikelihood} bounds the likelihood
through the condition number of the correlation matrix and argues that
likelihood-optimal models avoid arbitrarily ill-conditioned ones; here the
likelihood-optimal covariance is nonetheless more ill conditioned than
double precision resolves. In
radial basis function approximation, extended- and arbitrary-precision
studies place the error-optimal shape parameter beyond the reach of double
precision
\cite{Huang2007MultiquadricHighPrecision,Sarra2011ExtendedPrecisionRBF,Cheng2012MultiquadricShapeParameter},
but they compare a few fixed precisions and involve neither a
likelihood nor Cholesky failure. In numerical linear algebra, Cholesky
breakdown in low precision is avoided by scaling and shifting the
matrix~\cite{HighamPranesh2021LowerPrecisionSPD}, the counterpart of a
jitter. Low precision in GP training has been examined for large, noisy
problems with iterative
solvers~\cite{Gardner2018GPyTorch,Maddox2022LowPrecisionGP}, mixed-precision
tile Cholesky for geostatistical maximum likelihood treats double precision
as ground
truth~\cite{Abdulah2019MixedPrecisionCholesky,Abdulah2022MixedPrecisionPaRSEC},
and computation-aware GPs~\cite{Wenger2022PosteriorComputationalUncertainty}
model solver truncation, not rounding. Missing, to our knowledge, is a
\emph{quantitative} account, for small noiseless problems with smooth
kernels, of how \emph{maximum-likelihood} hyperparameters depend on working
precision, of how Cholesky failure displaces them, and of how many bits a
given kernel and design need.

\paragraph{Contributions}
\begin{enumerate}
  \item \emph{A precision sweep} of GP maximum likelihood over significand
    widths (\cref{sec:results:sweep}): for the squared-exponential kernel
    on a design with a BO-style cluster, the fits converge only once the
    width exceeds $\log_2\kappa_2(A_{\theta^\ast})$, the \emph{threshold},
    with $\kappa_2$ the spectral condition number and $\theta^\ast$ the
    exact MLE.
  \item \emph{An implicit-constraint mechanism, tested independently of the
    fits} (\cref{sec:theory:constraint,sec:results:factorability,sec:results:branch}).
    Cholesky failure confines the maximization to covariances with
    $\kappa_2(A_\theta)\lesssim1/u$ and displaces the computed MLE mostly
    along the weakly identified direction (a local expansion,
    \cref{prop:kkt}, is proved for the idealized problem); a factorability
    map, exact constrained solutions over a range of levels and a prediction from the working
    precision alone test this without the returned fits.
  \item \emph{Separation of arithmetic error from optimizer behavior}
    (\cref{sec:results:optimizers}). Below the threshold the handling of
    Cholesky failures, not the optimizer class, decides the outcome; above
    it a gradient polish attains hyperparameter errors linear in the
    arithmetic-error scale, against its square root for function-value
    methods.
  \item \emph{Generality, consequences and remedies}
    (\cref{sec:results:generality,sec:results:posterior,sec:results:remedies}).
    A benchmark with negative controls relates the bits
    needed to the conditioning at the reference optimum; consequences for
    the posterior and closed-loop active learning are measured, and
    remedies compared on fixed designs.
\end{enumerate}

\section{Background}\label{sec:background}

\paragraph{Type-II maximum likelihood}
Let $X=\{x_1,\dots,x_n\}\subset\mathbb{R}^d$ be the design and
$y\in\mathbb{R}^n$ the (standardized) observations. We use a zero-mean GP with
the anisotropic squared-exponential (ARD) kernel~\cite[Chaps.~4--5]{Rasmussen2006GPML}
\begin{equation}\label{eq:kernel}
  k_\theta(x,x') = \sigma_f^2 \exp\Bigl(-\tfrac12\sum_{j=1}^d
    \frac{(x_j-x'_j)^2}{\ell_j^2}\Bigr),
\end{equation}
where, in~\cref{eq:kernel} only, subscripts on $x$ and $x'$ denote
coordinates; elsewhere they index design points. The hyperparameters
$\theta=(\sigma_f,\ell_1,\dots,\ell_d,\sigma_n)$ are the signal standard
deviation, one length scale per input dimension, and the noise standard
deviation. With the kernel matrix
$K_\theta=[k_\theta(x_i,x_k)]_{i,k=1}^n$ and the covariance matrix
$A_\theta = K_\theta+\sigma_n^2 I$, the negative log-likelihood is
\begin{equation}\label{eq:nll}
  \mathrm{NLL}(\theta) = \tfrac12 y^\top A_\theta^{-1}y
    + \tfrac12\log\det A_\theta + \tfrac n2\log 2\pi,
\end{equation}
in nats, evaluated through the Cholesky factorization $A_\theta=LL^\top$.
The maximum-likelihood estimate (MLE) is
$\theta^\ast=\arg\min_\theta\mathrm{NLL}(\theta)$ over a box of admissible
values. Unadorned $\mathrm{NLL}$ and $\theta^\ast$ always refer to exact
arithmetic; computed counterparts carry a precision subscript
(\cref{sec:setup}).

\paragraph{Floating-point model}
A working precision is described by its significand width $b$ (including the
implicit bit), with unit roundoff $u=2^{-b}$: $b=11$, $24$, $53$ and
$64$ correspond to IEEE half, single, double and x87 extended precision
(the $80$-bit format).
We write $\mathrm{fl}(\cdot)$ for an expression evaluated at the working
precision and ``ulp'' for a unit in the last place; float64 denotes IEEE
double ($b=53$)~\cite{IEEE2019Std754}.

\paragraph{Cholesky in finite precision}
For symmetric positive definite $A$ let
$\kappa_2(A)=\lambda_{\max}(A)/\lambda_{\min}(A)$ be the spectral condition
number and $R=D^{-1}AD^{-1}$, with $D=\operatorname{diag}(A)^{1/2}$, the
unit-diagonal scaling of $A$. Demmel's
criterion~\cite[Thm.~3.2]{Demmel1989CholeskyErrors} gives a sufficient
condition for success and a partial converse:
if $\lambda_{\min}(R)>(n^2+n)u/(1-(n-1)u)$ and no underflow occurs,
Cholesky applied to $A$ succeeds, whereas if $\lambda_{\min}(R)\le u$ there
exist rounding errors that make it fail. Throughout, $a\lesssim c$ means
that $a$ is at most an untracked constant, independent of the precision $u$,
times $c$, and $a\asymp c$
means both $a\lesssim c$ and $c\lesssim a$.

\paragraph{Conditioning of Gaussian kernel matrices}
Gaussian kernel matrices become ill conditioned as points cluster, as in BO
designs where a converging optimizer piles points up near the
incumbent~\cite{Mohammadi2016AnalyticRegularization,McLeod2018OptimizationFastSlow},
and as the length scales grow relative to the design spacing (the flat
limit)~\cite{FornbergWrightLarsson2004FlatRBF,Schaback2005PolynomialsRBF};
by Schaback's trade-off principle~\cite{Schaback1995ErrorConditionNumbers},
accurate interpolants and ill-conditioned matrices go together. Write
$K_{t\ell}$ for the kernel matrix with every length scale $\ell_j$ replaced
by $t\ell_j$ ($t>0$, other hyperparameters fixed), $\Pi_\nu^d$ for the
polynomials in $d$ variables of total degree at most $\nu$, a space of
dimension $\binom{\nu+d}{d}$, and $\nu(X)$ for the smallest $\nu$ such that
$\Pi_\nu^d$ restricted to $X$ spans $\mathbb{R}^n$; for unisolvent $X$,
which holds almost surely for points drawn from a density, $\nu(X)$ is the
smallest degree with $\dim\Pi_\nu^d\ge n$
\cite[Thms.~6.1--6.2]{BarthelmeUsevich2021FlatLimit}. As $t\to\infty$ the
eigenvalues of a Gaussian $K_{t\ell}$ separate into groups of exact orders
$t^0,t^{-2},\dots,t^{-2\nu(X)}$~\cite[Thm.~6 and its
proof]{Schaback2005PolynomialsRBF} (see also~\cite{BarthelmeUsevich2021FlatLimit}), so
$\kappa_2(K_{t\ell})\asymp t^{2\nu(X)}$ with constants depending on $X$ and
$\ell$; anisotropic $\ell$ reduces to the isotropic case by rescaling the
coordinates, which leaves $\nu(X)$ unchanged.

\section{Experimental setup}\label{sec:setup}

\subsection{Test problem}
A zero-mean GP with the ARD kernel~\cref{eq:kernel} in $d=2$ emulates the
Branin function~\cite{Branin1972WidelyConvergent} rescaled to $[0,1]^2$,
with standardized targets. The design has $n=30$ points: $20$ scrambled
Sobol points~\cite{Sobol1967DistributionPoints,Owen1998ScramblingSobol} plus $10$ points of the kind
a converging BO run adds around a minimizer (random seed 3; minimum
pairwise distance about $1.5\times10^{-3}$; \cref{app:protocols}). No jitter
is added; the only diagonal regularization is $\sigma_n^2$. The box constraints are
$\sigma_f\in[0.05,20]$, $\ell_j\in[0.02,10]$ and
$\sigma_n\in[10^{-9},0.5]$.

\subsection{Precision model}
All arithmetic in the likelihood (kernel evaluation, Cholesky factorization,
triangular solves) and in prediction is performed at $b$ significand bits,
for $b$ from $11$ to $256$ (\cref{tab:sweep,app:protocols}), in
mpmath~\cite{Mpmath2026} with an effectively unbounded exponent, so only
significand rounding is simulated; overflow, underflow, and IEEE fused
operations are not. The training inputs and targets are held at the
reference precision, so only the arithmetic is rounded. A Cholesky failure
(a non-positive pivot) returns a large penalty to the optimizer.

\subsection{Optimizer and protocols}
The optimizer is Powell's method~\cite{Powell1964EfficientMinimization} on
the logarithms of the hyperparameters, ported from
SciPy~\cite{Virtanen2020SciPy} to mpmath and run in the same $b$-bit arithmetic as the
objective, from two fixed starts (\cref{app:protocols}). The fitted vector
$\theta_b$ is the returned point with the lower $b$-bit NLL of the two
starts. The \emph{independent} protocol uses the two shared
starts at every precision; the \emph{continuation} protocol additionally
warm-starts from the optimum at the previous precision.

A projected Newton method~\cite{Bertsekas1982ProjectedNewton} with the
analytic gradient and Hessian
of~\cref{eq:nll} polishes fits at their own working precision
(\cref{app:protocols}). The reference $\theta_{\mathrm{ref}}$, our proxy for
$\theta^\ast$, is the $384$-bit Powell fit after this polish. We chose $384$ bits, halfway
between $256$ and $512$, to lie well above both the largest tested precision
and $\log_2\kappa_2(A_{\theta_{\mathrm{ref}}})\approx70$; a Newton polish at
$512$ bits moves $\log\theta_{\mathrm{ref}}$ by at most
$1.1\times10^{-83}$.
The tables
and figures use the Powell fits unless stated.

\subsection{Metrics}\label{sec:setup:metrics}
We write $\mathrm{NLL}_b(\theta)$ for the NLL~\cref{eq:nll} computed in
$b$-bit arithmetic, $\mathrm{NLL}_{\mathrm{ref}}=\mathrm{NLL}_{384}$, and
$\theta_{b,i}$ for the $i$th component of $\theta_b$ (logarithms act
componentwise). Every fitted $\theta_b$ is judged at the reference
precision by (i) the hyperparameter error
$\Delta=\max_i\lvert\log\theta_{b,i}-\log\theta_{\mathrm{ref},i}\rvert$; (ii) the
likelihood loss
$\mathrm{NLL}_{\mathrm{ref}}(\theta_b)-\mathrm{NLL}_{\mathrm{ref}}(\theta_{\mathrm{ref}})\ge0$;
(iii) the relative $L^2$ change in the posterior mean on a $31\times31$ grid
over the domain due to the hyperparameters alone, both posteriors being
evaluated at $384$ bits on the same data; and (iv) the objective error at the
reference point,
$\varepsilon_f(b)=\lvert\mathrm{NLL}_b(\theta_{\mathrm{ref}})-\mathrm{NLL}_{\mathrm{ref}}(\theta_{\mathrm{ref}})\rvert$,
defined only when the $b$-bit Cholesky succeeds at $\theta_{\mathrm{ref}}$.

We write $H$ for the NLL Hessian at $\theta_{\mathrm{ref}}$, computed at
$384$ bits by central differences (step $10^{-3}$) in the free
log-parameters $(\log\sigma_f,\log\ell_1,\log\ell_2)$, since $\sigma_n$ sits
on its lower bound. Its eigenvalues are
$\lambda_1\le\lambda_2\le\lambda_3$ with unit eigenvectors $v_1,v_2,v_3$, so
$v_1$ is the flattest direction. Code and raw results will be released at
\url{https://github.com/tscott-00/gpr_precision}; the repository stays
empty until publication.

\subsection{Further experiments}\label{sec:setup:followup}
Four further experiments reuse the data, bounds, starts and judging
precision above unless stated (protocols in \cref{app:protocols}). A
\emph{factorability map} evaluates $b$-bit Cholesky at fixed points on both
sides of the boundary, with splits separating entry rounding from the
factorization, and the constrained problems of
\cref{sec:theory:constraint} are solved as $384$-bit KKT points over a
range of constraint levels. We use \emph{branch} to mean the curve of
solutions traced out as the level varies, with two cases: the
\emph{fixed-noise} branch, with $\sigma_n$ held at its floor, and the
\emph{free} branch, with $\sigma_n$ free
(\cref{sec:results:factorability,sec:results:branch}). An
\emph{optimizer control} repeats the fit near double precision from shared
starts with methods that differ in class (Powell, L-BFGS~\cite{Byrd1995LimitedMemoryBound,Zhu1997Algorithm778LBFGSB}, Newton) and in
whether a Cholesky failure returns a zero gradient or is rejected by the
line search (\cref{sec:results:optimizers}). A \emph{generality benchmark}
measures thresholds on Newton-polished fits for the squared-exponential
(SE), Mat\'ern $3/2$ and Mat\'ern
$5/2$ kernels~\cite{Stein1999InterpolationSpatialData,Rasmussen2006GPML} on clustered designs like
the one above and on Sobol designs over several seeds (Stage A), and for
variations of one ingredient (cluster geometry, size, noise, dimension) of
the seed-$3$ SE problem (Stage B; \cref{sec:results:generality}).
\emph{Remedies} (a jitter $j$, that is a fixed variance added to the
diagonal of $A_\theta$; extended precision; a high-precision refit; and a
library) are compared on the test problem and the final active-learning
designs of \cref{sec:results:bo}, with the same float64 Powell outer
optimizer in every arm but the library's (\cref{sec:results:remedies}).

\section{Mechanisms}\label{sec:theory}

\Cref{prop:kkt} is proved (\cref{app:kkt}) for an idealized problem,
\cref{hyp:constraint} is an empirical model tested in
\cref{sec:results:factorability,sec:results:prediction}, and the estimates
of \cref{sec:theory:resolution} are first-order heuristics whose constants
are measured, not derived.

\subsection{Cholesky failure as an implicit constraint}\label{sec:theory:constraint}
Let $R_\theta=D^{-1}A_\theta D^{-1}$ with
$D=\operatorname{diag}(A_\theta)^{1/2}$, and write
$g(\theta)=\log_2\kappa_2(A_\theta)$ and
$h(\theta)=-\log_2\lambda_{\min}(R_\theta)$; for stationary kernels $D$ is
a multiple of $I$ and $g(\theta)-h(\theta)=\log_2\lambda_{\max}(R_\theta)$,
which lies in $[0,\log_2n]$ because $\operatorname{tr}R_\theta=n$. In terms
of the margin $x=\log_2(\lambda_{\min}(R_\theta)/u)=b-h(\theta)$ in bits,
Demmel's criterion (\cref{sec:background}) guarantees that $b$-bit Cholesky
succeeds for $x\gtrsim\log_2(n^2+n)$ and admits failing roundings for
$x\le0$, at every precision, so factorability data from all $b$ can be
pooled on one axis. It does not say where deterministic rounding actually
fails; that is measured against $x$ in \cref{sec:results:factorability},
and leads to the following working model.

\begin{hypothesis}[implicit constraint]\label{hyp:constraint}
Cholesky factorization in $b$-bit arithmetic succeeds whenever
$\lambda_{\min}(R_\theta)\ge c_2u$, that is $x\ge\log_2c_2$, with $c_2$ of order one, and below that edge success and failure depend on the
particular rounding errors over a band several bits wide. A computed
minimizer therefore lies in the factorable set, and if the exact MLE lies
outside it and there is no interior stationary point, on the edge of that
set.
\end{hypothesis}

Because penalized evaluations are discarded by the optimizer, the computed
problem is effectively the minimization of $\mathrm{NLL}(\theta)$ subject to
$h(\theta)\le b+\beta$, that is $x\ge-\beta$, with an offset
$\beta=O(1)$ in bits and a fuzzy boundary. The choice $\beta=-\log_2c_2$ is
the safe edge of \cref{hyp:constraint} ($\beta_{\mathrm{strict}}$ in
\cref{sec:results:prediction}), and larger $\beta$ moves the level into the
band where success depends on the rounding. Since
$g(\theta)-h(\theta)=\log_2\lambda_{\max}(R_\theta)$ varies far less than
$g(\theta)$ over the sampled points (\cref{app:values}), the same
constraint can be written in terms of the condition number,
\begin{equation}\label{eq:constrained}
  \min_\theta\ \mathrm{NLL}(\theta)\quad\text{s.t.}\quad g(\theta)\le b+\beta_g,
\end{equation}
with $\beta_g\approx\beta+\log_2\lambda_{\max}(R_\theta)$. \Cref{prop:kkt}
and the branches of \cref{sec:results:branch} use this form; the
predictions of \cref{sec:results:prediction} use the $h$ form and read
$\beta$ off the measured edge.

\begin{proposition}[displacement under a binding constraint]\label{prop:kkt}
Let $\theta$ denote the free log-parameters (any other hyperparameters
held fixed), and let $\mathrm{NLL}(\theta)$ and $g(\theta)$ be three times
continuously differentiable on an open set $U\subseteq\mathbb{R}^p$. Let $\theta^\ast\in U$ be a nondegenerate MLE,
$\nabla\mathrm{NLL}(\theta^\ast)=0$ and
$H=\nabla^2\mathrm{NLL}(\theta^\ast)\succ0$, with $s=\nabla g(\theta^\ast)\ne0$,
and let $q=s^\top H^{-1}s$. Let $r_b=g(\theta^\ast)-(b+\beta_g)$ be the
violation of the constraint at $\theta^\ast$ in bits, with the level
$b+\beta_g$ regarded as a continuous parameter. Then there are a
neighborhood $N$ of $\theta^\ast$ and $r_0>0$ such that, for $0<r_b<r_0$,
problem~\cref{eq:constrained} restricted to $N$ has a unique minimizer
$\theta^{\mathrm{c}}_b$, the constraint is active there, and, as $r_b\to0^+$,
\begin{equation}\label{eq:kktloss}
  \theta^{\mathrm{c}}_b=\theta^\ast-\frac{r_b}q\,H^{-1}s+O(r_b^2),\qquad
  \mathrm{NLL}(\theta^{\mathrm{c}}_b)-\mathrm{NLL}(\theta^\ast)=\frac{r_b^2}{2q}+O(r_b^3).
\end{equation}
\end{proposition}

The leading terms solve the quadratic model
$\min_{\delta\theta}\tfrac12\delta\theta^\top H\,\delta\theta$ subject to $s^\top\delta\theta\le-r_b$, and
the multiplier $\mu=r_b/q+O(r_b^2)$ is the derivative of the optimal value with
respect to the violation $r_b$ (minus its derivative with respect to the
level), the standard sensitivity of a constrained
minimizer~\cite{Nocedal2006NumericalOptimization}; \cref{app:kkt} gives a
self-contained proof.
We write $\theta^{\mathrm{c}}_b$ for constrained solutions to keep them
apart from the computed fits $\theta_b$. At
$\theta_{\mathrm{ref}}$ the NLL Hessian is only moderately conditioned,
whereas the covariance is badly conditioned (\cref{app:values}). The
displacement $H^{-1}s$ is largest along low-curvature directions aligned
with $s$, here the flattest eigenvector $v_1$, the weakly identified
$\sigma_f$--$\ell_2$ ridge: weak identifiability (cf.\
\cite{Zhang2004InconsistentEstimation,Stein1999InterpolationSpatialData})
amplifies precision bias without causing it.

The proposition concerns the sharp-constraint problem~\cref{eq:constrained};
that computed fits solve it is the content of \cref{hyp:constraint}. Its
smoothness hypothesis on $g(\theta)$ holds when the extreme eigenvalues of $A_\theta$
are simple (\cref{app:kkt}), which we assume at $\theta_{\mathrm{ref}}$ but
have not checked. The proposition also holds $\sigma_n$ at its floor; since raising
$\sigma_n$ increases the NLL but relieves the constraint, $\sigma_n$ leaves
its bound once $\mu$ exceeds the ratio of the slopes of $\mathrm{NLL}(\theta)$ and
$-g(\theta)$ in $\log\sigma_n$. We call the part of the free branch beyond this
point, where $\sigma_n$ is above its floor, the \emph{released} branch; the
sweep fits below the threshold have $\sigma_n$ above its floor, as on that
branch (\cref{sec:results:branch}). Once $\theta_{\mathrm{ref}}$ is feasible,
optimizer resolution sets the loss (\cref{sec:theory:resolution}); at low
precision the local expansion fails and an implicit nugget takes over
(\cref{sec:results:regimes}).

\subsection{Bits needed once the reference is factorable}\label{sec:theory:resolution}\label{sec:theory:objective}\label{sec:theory:flat}
\paragraph{Objective error}
If the computed $A_\theta$ and its factorization are exact for $A+\delta A$
with $\lVert\delta A\rVert_2\le\eta\lVert A\rVert_2$ and $\eta=c(n)u$, where
$c(n)$ is a low-degree polynomial in $n$, independent of $A$ and $u$, as the
backward error analysis of Cholesky~\cite[Chap.~10]{Higham2002Accuracy}
provides for the factorization,
then to first order the log-determinant changes by
$\operatorname{tr}(A^{-1}\delta A)$ and $y^\top A^{-1}y$ by
$-\alpha^\top\delta A\,\alpha$, $\alpha=A^{-1}y$, so
\begin{equation}\label{eq:objerr}
  \lvert\mathrm{NLL}_b-\mathrm{NLL}\rvert\lesssim\tfrac12\eta\kappa_2(A)\,
    \bigl(n+y^\top A^{-1}y\bigr)\qquad\text{for }\eta\kappa_2(A)<1,
\end{equation}
the $\kappa u$ noise model that Basak et
al.~\cite{Basak2022NumericalIssuesMLE} use in double precision. We use only
its scaling in $\kappa_2u$, also outside the regime $\eta\kappa_2<1$ where
it is derived; the constant is measured in \cref{sec:results:optimizers},
and worst-case rounding constants of this kind are known to be
pessimistic~\cite{HighamMary2019ProbabilisticRounding}.

\paragraph{Optimizer resolution}
With objective error at most $\varepsilon_f$, a function-value method cannot
distinguish displacements $\delta\theta=\theta-\theta^\ast$ with
$\tfrac12\delta\theta^\top H\,\delta\theta\lesssim2\varepsilon_f$
\cite[\S8.2.2]{Gill2019PracticalOptimization}; a Newton iteration with
gradient error $\varepsilon_g$ sees the displacement linearly,
$\nabla\mathrm{NLL}(\theta)\approx H\,\delta\theta$. Along a unit eigenvector $v_i$ of
$H=\nabla^2\mathrm{NLL}(\theta^\ast)$ with eigenvalue $\lambda_i$ (computed
at $\theta_{\mathrm{ref}}$, our proxy for $\theta^\ast$;
\cref{sec:setup:metrics}) the attainable accuracies are therefore
\begin{align}
  \lvert v_i^\top\delta\theta\rvert&\lesssim2\sqrt{\varepsilon_f/\lambda_i}
    &&\text{(function values)},\label{eq:resolution}\\
  \lvert v_i^\top\delta\theta\rvert&\lesssim\lVert\varepsilon_g\rVert/\lambda_i
    &&\text{(gradients)},\label{eq:gradresolution}
\end{align}
so when $\varepsilon_g$ and $\varepsilon_f$ are comparable and small the
gradient bound is roughly the square of the function-value bound, hence much
smaller, and the optimizer class matters. Combining
\cref{eq:resolution} with~\cref{eq:objerr}, recovering $\theta^\ast$ to a
tolerance $\tau$ along $v_1$ needs roughly
$b\gtrsim\log_2\kappa_2+\log_2\bigl(C/(\lambda_1\tau^2)\bigr)$ bits, with
$\kappa_2=\kappa_2(A_{\theta^\ast})$ and $C$ collecting the constants; with
gradient resolution~\cref{eq:gradresolution} $\tau^2$ becomes $\tau$. Rule
R2 of \cref{sec:results:generality} fits $C$ for each tolerance separately,
so the power of $\tau$ does not affect it.

\paragraph{Flat-limit growth}
Since $\kappa_2(K_{t\ell})\asymp t^{2\nu(X)}$ as $t\to\infty$
(\cref{sec:background}), the bits needed grow as $2\nu(X)\log_2t+O(1)$
under a uniform dilation of the length scales. For unisolvent $X$ with
$n=30$ and $d=2$, $\dim\Pi_6^2=28<30\le36=\dim\Pi_7^2$, so $\nu(X)=7$ and the
predicted growth is $14$ bits per doubling; $\sigma_n^2$ caps the growth
for $A$. \Cref{app:values} compares this with the growth measured from
$\theta_{\mathrm{ref}}$ outward. The prediction depends on smoothness
relative to the design, not on the closest pair of points that
separation-based bounds~\cite{NarcowichWard1992NormEstimates,Schaback1995ErrorConditionNumbers}
use.

\section{Results}\label{sec:results}

\Cref{sec:results:sweep} measures the fits on the test problem of
\cref{sec:setup}, and
\cref{sec:results:factorability,sec:results:branch,sec:results:optimizers}
test their mechanism independently of the returned fits;
\cref{sec:results:generality,sec:results:posterior,sec:results:remedies}
turn to other kernels, designs and seeds, to the posterior and active
learning, and to remedies.

\subsection{The precision sweep}\label{sec:results:sweep}\label{sec:results:regimes}
Double precision returns a different model, not a perturbed one: the
$53$-bit fit loses about $13$ nats at the reference precision, and a loss
remains at $64$ bits (\cref{tab:sweep}). Fits converge only once $b$
exceeds the threshold $\log_2\kappa_2(A_{\theta_{\mathrm{ref}}})\approx70$.
Below it more bits do not reliably help: the error is not monotone in $b$.
From $72$ bits no
evaluation fails and $\Delta$ drops abruptly, to a common floor from $128$
bits that is set by Powell's stopping test, not by the arithmetic
(\cref{sec:results:optimizers}).
Warm starts from the previous precision
(the continuation protocol) still leave losses of several nats at $53$ and
$56$ bits (\cref{tab:optimizers}(a)), so the computed objective, not only the
starting point, limits the fit.

\begin{table}[tbhp]
\centering
\caption{Independent-protocol fits against significand bits $b$, judged at
$384$ bits (\cref{sec:setup}): failed Cholesky evaluations,
$\Delta=\max_i\lvert\log\theta_{b,i}-\log\theta_{\mathrm{ref},i}\rvert$,
likelihood loss (nats), relative $L^2$ change of the posterior mean, and
$g(\theta_b)-b$ with $g=\log_2\kappa_2(A_\theta)$ computed at $384$ bits.
$^\dagger$: at a bound. The $384$-bit reference agrees with the last row in
the four hyperparameter columns.}
\label{tab:sweep}
\footnotesize
\setlength{\tabcolsep}{2.5pt}
\begin{tabular}{rllll rllll}
\toprule
$b$ & $\sigma_f$ & $\ell_1$ & $\ell_2$ & $\sigma_n$ & fail & $\Delta$ & loss & mean & $g(\theta_b)-b$ \\
\midrule
16 & $2.02$ & $0.230$ & $0.230$ & $0.0137$ & 163 & $16.4$ & $73.7$ & $0.441$ & $2.1$ \\
24 & $1.08$ & $0.225$ & $0.302$ & $5.39\times10^{-4}$ & 119 & $13.2$ & $45.7$ & $0.364$ & $1.7$ \\
32 & $1.74$ & $0.221$ & $0.545$ & $3.91\times10^{-5}$ & 93 & $10.6$ & $29.4$ & $0.144$ & $2.8$ \\
40 & $17.9$ & $0.302$ & $2.31$ & $3.00\times10^{-5}$ & 38 & $10.3$ & $22.3$ & $0.0455$ & $2.7$ \\
48 & $2.42$ & $0.246$ & $0.924$ & $2.46\times10^{-7}$ & 87 & $5.51$ & $16.4$ & $0.0563$ & $2.6$ \\
53 & $1.79$ & $0.228$ & $0.549$ & $3.72\times10^{-9}$ & 180 & $1.98$ & $13.3$ & $0.108$ & $4.2$ \\
56 & $1.53$ & $0.221$ & $0.604$ & $4.62\times10^{-9}$ & 31 & $2.14$ & $15.2$ & $0.105$ & $1.7$ \\
60 & $5.70$ & $0.272$ & $1.16$ & $6.14\times10^{-9}$ & 65 & $1.81$ & $1.32$ & $0.0284$ & $3.5$ \\
64 & $9.76$ & $0.293$ & $1.52$ & $1.94\times10^{-9}$ & 6 & $0.663$ & $0.202$ & $8.26\times10^{-3}$ & $3.6$ \\
72 & $12.7$ & $0.304$ & $1.76$ & $1.02\times10^{-9}$ & 0 & $0.0172$ & $3.89\times10^{-3}$ & $2.90\times10^{-4}$ & $-2.5$ \\
80 & $12.9$ & $0.304$ & $1.78$ & $1.00\times10^{-9}$ & 0 & $5.77\times10^{-4}$ & $4.96\times10^{-5}$ & $1.72\times10^{-5}$ & $-10.4$ \\
96 & $12.9$ & $0.304$ & $1.78$ & $1.00\times10^{-9}$$^\dagger$ & 0 & $2.66\times10^{-3}$ & $4.64\times10^{-5}$ & $3.37\times10^{-5}$ & $-26.4$ \\
128--256 & $12.9$ & $0.304$ & $1.78$ & $1.00\times10^{-9}$$^\dagger$ & 0 & $4.85\times10^{-7}$ & $7.33\times10^{-11}$ & $8.6\times10^{-9}$ & $69.6-b$ \\
\bottomrule
\end{tabular}
\end{table}

For $16\le b\le64$ the excess $g(\theta_b)-b$ stays roughly constant
(\cref{tab:sweep}; both protocols in \cref{app:values}): the fits sit on the edge of the factorable set and buy
factorability in one of two ways. For $b\le48$ the optimizer raises the
noise, an implicit nugget with $\sigma_n^2\approx3u\sigma_f^2$ that alone
sets $\lambda_{\min}(R_{\theta_b})$ (two continuation fits excepted;
\cref{app:values}), the precision-limited analogue of choosing the smallest
nugget that makes $\kappa$ acceptable~\cite{Ranjan2011ComputationallyStable}.
Between about $53$ and $64$ bits the noise is close to its floor, and the
fit instead retreats along the weakly identified $\sigma_f$--$\ell_2$
ridge. Once $\sigma_n^2<u\sigma_f^2/2$,
$\mathrm{fl}(\sigma_f^2+\sigma_n^2)=\sigma_f^2$ on the diagonal and the
computed NLL does not depend on $\sigma_n$ at all.

\subsection{The factorable set, measured independently}\label{sec:results:factorability}
\begin{figure}[tbhp]
\centering
\includegraphics[width=\textwidth]{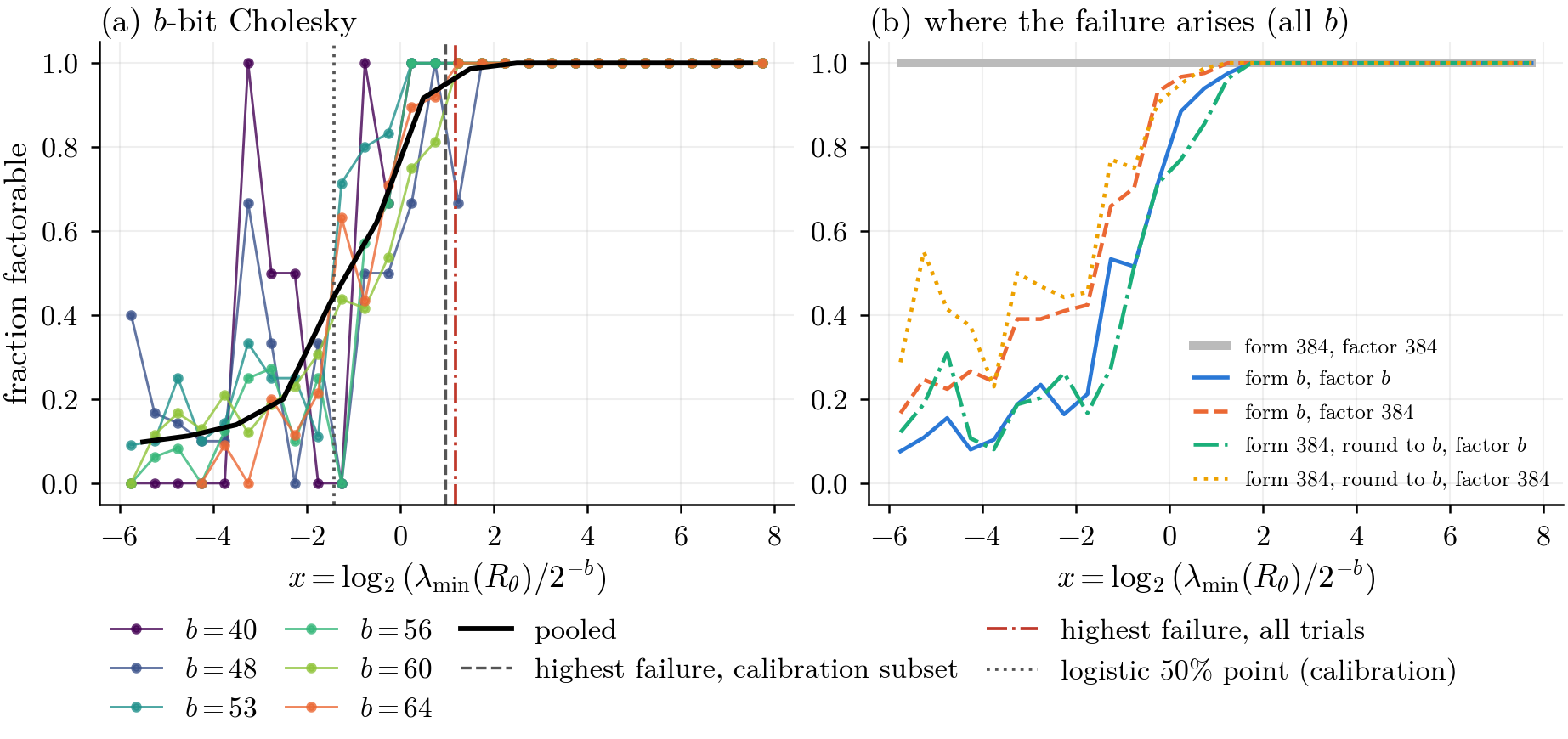}
\caption{(a) Fraction of trials whose $b$-bit Cholesky
factorization succeeds, against $x=\log_2(\lambda_{\min}(R_\theta)/u)$ binned
over $-6\le x<8$: per precision for $b\le64$ (every point factors from
$68$ bits) and pooled over $b=40$--$72$ (black). Dashed: largest $x$ with a failure in the
calibration subset ($0.97$), which sets the offsets of
\cref{fig:prediction}; red dash-dotted: over all trials ($1.19$, a held-out
$48$-bit point); dotted: logistic $50\%$ point of the calibration
subset ($-1.42$). (b) The same fraction
for the arithmetic splits of \cref{app:protocols}; gray: entries formed and
factored at $384$ bits, which succeeds at every point.}
\label{fig:factorability}
\end{figure}

The factorability map (protocol in \cref{app:protocols}) measures factorability directly, at
fixed parameter points rather than at points an optimizer returned
(\cref{fig:factorability}). The edge is calibrated on a subset of the
points and checked on the rest (\cref{app:protocols}). On the calibration
subset the largest $x$ with a failure is $0.97$, that is
$\lambda_{\min}(R_\theta)=2u$ and $c_2\approx2$ in \cref{hyp:constraint},
and every one-bit bin from $x=1$ upward factors in at least $95\%$ of
trials, the $95\%$ edge. Over all trials the only failures above $0.97$
are two held-out points, the $48$-bit constrained solutions at the offset
$\beta_{95}=-1$ of that edge, at $x=1.00$ and $1.19$; every matrix with
$\lambda_{\min}(R_\theta)\ge2.3u$ ($x>1.19$) factors. The calibrated edge
is thus the largest failing $x$ in the calibration subset, not a guarantee,
and the boundary stays fuzzy by about $0.2$ bits beyond it. All offsets
used below are calibrated on this subset, not on the full map. Below that
edge success depends on the particular rounding over a band about ten
bits wide, with isolated successes far below it and no trend in $b$; the
$h$ and $g$ descriptions of the boundary are equivalent here
(\cref{app:values}). These fractions are over sampled matrices with
deterministic rounding, not probabilities.

The constraint belongs to the stored matrix rather than to the
factorization: most points that fail the full $b$-bit pipeline already fail
when entries formed at $384$ bits are rounded once to $b$ bits and factored
at $384$ bits (\cref{fig:factorability}(b)). \Cref{app:values} gives the
fractions per split, the failing pivots and the NLL error of points that do
factor.

\subsection{Constrained solutions}\label{sec:results:branch}\label{sec:results:prediction}
\begin{figure}[tbhp]
\centering
\includegraphics[width=\textwidth]{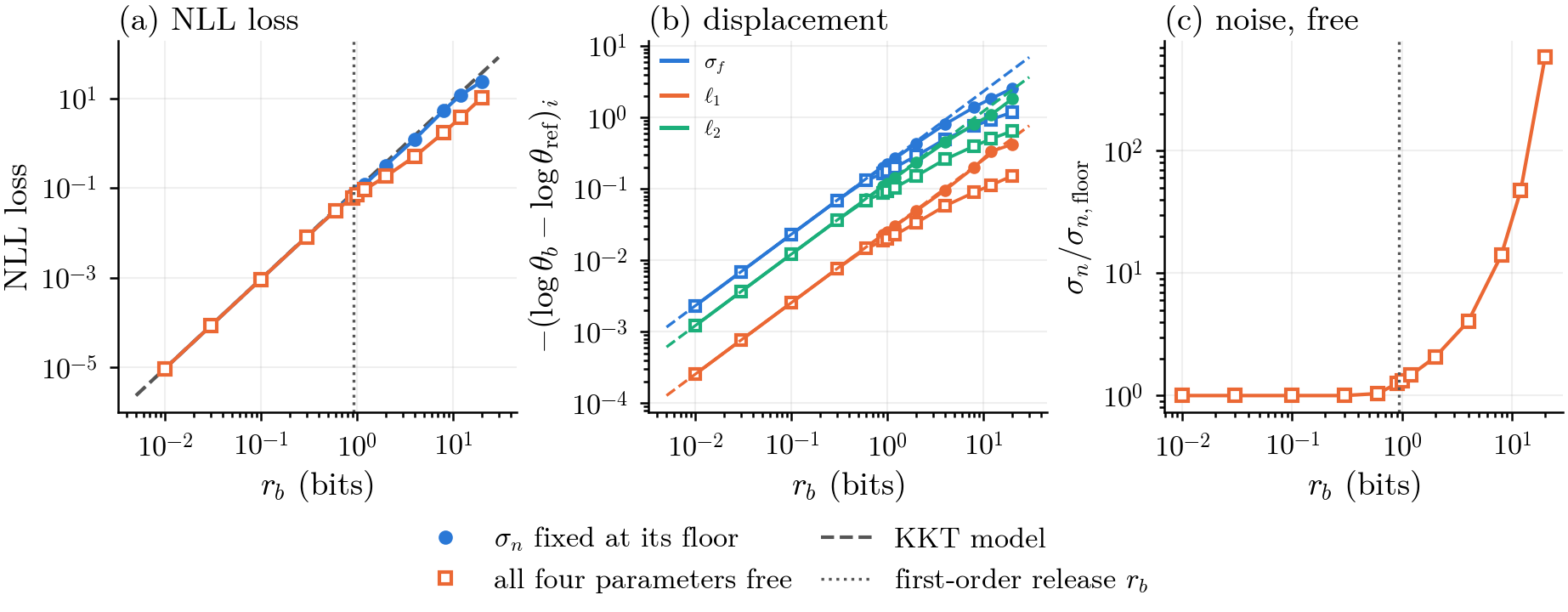}
\caption{Constrained MLE at the level $g(\theta)\le g(\theta_{\mathrm{ref}})-r_b$
with $\sigma_n$ at its floor (dots) and free (squares). (a) Loss and the
leading term $r_b^2/(2q)$, $q=s^\top H^{-1}s$, of \cref{prop:kkt} (dashed). (b) Displacement
$-(\log\theta^{\mathrm{c}}_b-\log\theta_{\mathrm{ref}})_i$ in $(\sigma_f,\ell_1,\ell_2)$
and $r_bH^{-1}s/q$ (dashed); $\theta^{\mathrm{c}}_b$: constrained
solution. (c) Free-branch $\sigma_n$ over its floor.
Dotted: first-order release estimate $r_b=0.93$.}
\label{fig:branch}
\end{figure}

\Cref{fig:branch} follows the constrained MLE~\cref{eq:constrained} as the
level is lowered to $g(\theta_{\mathrm{ref}})-r_b$, where $r_b$ is the
violation of \cref{prop:kkt} at $\theta^\ast=\theta_{\mathrm{ref}}$; the
solver follows $r_b$ directly, as if the level $b+\beta_g$ were continuous. Like $g(\theta)$, $r_b$ is measured
in bits, that is, in $\log_2$ units of $\kappa_2$, so it takes continuous
values: the constraint requires $\kappa_2$ to fall by a factor $2^{r_b}$ from its
value at $\theta_{\mathrm{ref}}$. The solutions as $r_b$ varies trace out two
branches, the fixed-noise
branch with $\sigma_n$ held at its floor and the free branch with all four
parameters free. Each point is a $384$-bit KKT point (\cref{app:protocols}): the
best of a finite multistart, not certified globally optimal.
\Cref{prop:kkt} describes the fixed-noise branch at small violations;
that branch has the predicted direction and size of displacement, but its
loss follows the leading term of~\cref{eq:kktloss} closely only within a
fraction of a bit of violation (\cref{tab:branch}). The free branch follows
the fixed-noise branch until $\sigma_n$ leaves its floor, earlier than the
first-order release estimate (dotted in \cref{fig:branch}); beyond that
the noise grows by orders of magnitude and the loss falls well below the
fixed-noise loss. Every sweep fit below $72$ bits has $g(\theta_b)$ about
two bits or more below $g(\theta_{\mathrm{ref}})$ and $\sigma_n$ above its
floor (\cref{tab:sweep}), as on the released branch. From $48$ to $64$ bits
the fixed-noise model gives the direction of their displacement but
over-predicts its size and the loss; lower-precision fits are displaced
less, and the $40$-bit fit in a different direction (\cref{app:values}).

\begin{figure}[tbhp]
\centering
\includegraphics[width=\textwidth]{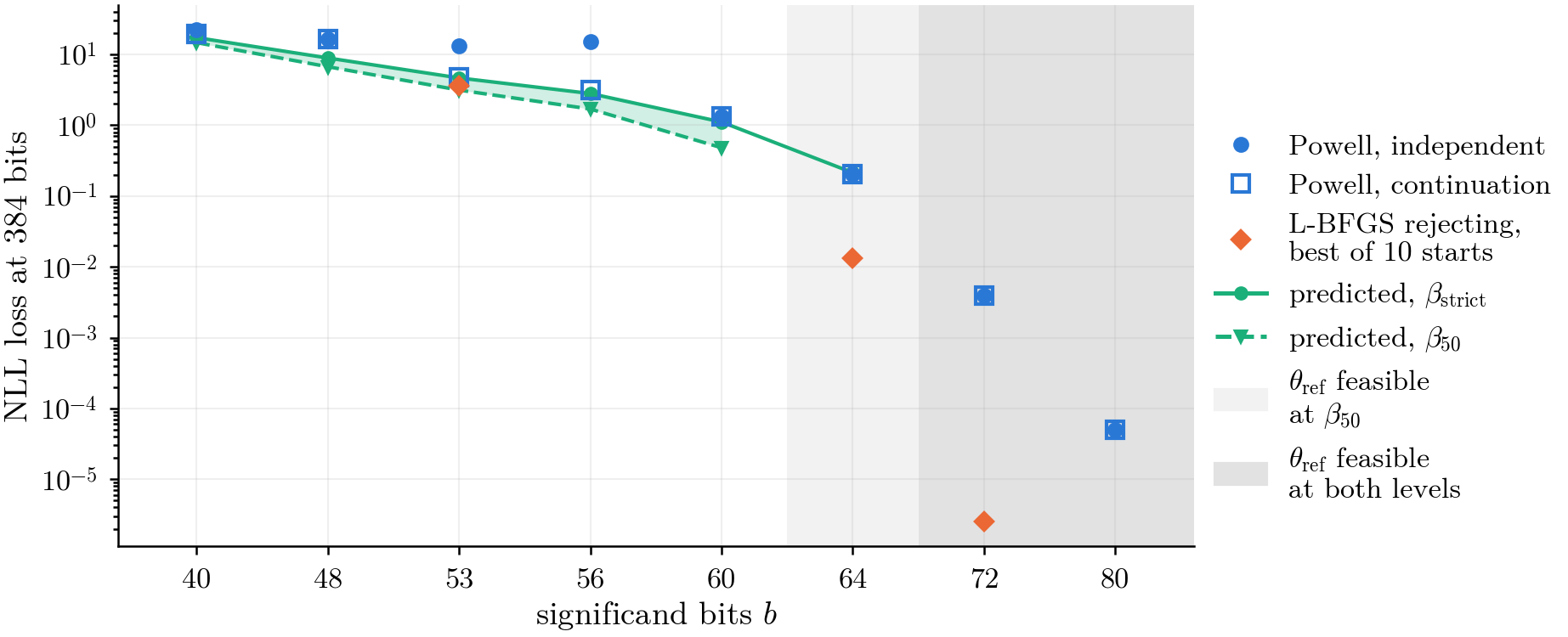}
\caption{Loss at $384$ bits of the sweep's Powell fits and of the best of
ten rejecting-L-BFGS starts (\cref{tab:optimizers}(b)), against the
constrained solution predicted from $b$ alone
($\min\mathrm{NLL}$ s.t.\ $h(\theta)\le b+\beta$, $\sigma_n$ free) at
$\beta_{\mathrm{strict}}=-0.97$ (solid) and $\beta_{50}=1.42$ (dashed),
shaded between them where both bind. Gray columns: $\theta_{\mathrm{ref}}$
is feasible at $\beta_{50}$ (light) or at both levels (darker), so the
predicted loss is zero.}
\label{fig:prediction}
\end{figure}

The constraint model also predicts the fits themselves, from $b$ alone and
with nothing fitted to them. An offset $\beta$ read off the factorability
map turns the edge $x=b-h\ge x_0$ of \cref{fig:factorability}(a) into the
level $h(\theta)\le b+\beta$ with $\beta=-x_0$; we solve the free-noise
constrained problem at that level and compare its loss with the fits'
(\cref{fig:prediction}; \cref{tab:optimizers}(a)). The strict offset
$\beta_{\mathrm{strict}}=-0.97$ comes from the largest $x$ with a failure
in the calibration subset (dashed in \cref{fig:factorability}(a)), not from
the full map ($1.19$), and $\beta_{50}=1.42$ from the subset's logistic
$50\%$ point (dotted). The predicted solutions at
$\beta_{\mathrm{strict}}$ factor in $b$-bit arithmetic at every $b$, those
at $\beta_{95}$ fail at $48$ bits, and at $\beta_{50}$, and at looser
offsets up to the lowest $x$ at which any sampled matrix factors, they fail
for $b\le56$. Using $g$ in place of $h$ changes the predicted losses by at
most $4\%$ except at $\beta_{95}$ (\cref{app:values}).

From $40$ to $64$ bits the best Powell fit at each precision is within
$20\%$ of the strict prediction, with one exception: at $48$ bits both
protocols return the same fit, at about twice the predicted loss. Like the
failures at $53$ and $56$ bits below, this is an optimizer failure, not a
tighter factorable set: the fit lies about
half a bit inside the edge ($x=1.54$), with a smaller $\sigma_f$ than the
prediction, and the predicted point factors in $48$-bit arithmetic with a
$48$-bit computed NLL $6.4$ nats below the fit's. The optimizer control,
run only at $53$, $64$ and $72$ bits, does not test this directly.
The independent-protocol fits at $53$ and $56$ bits lie several times above
the prediction, while the continuation fits at the same precisions reach
it, so these are optimizer failures, not a property of the precision. At
$53$ bits the best rejecting-L-BFGS start lands between the losses
predicted at the two levels, as a fuzzy boundary allows.

A feasible $\theta_{\mathrm{ref}}$ means only that factorability no longer
forces the fit away from it; two other limits remain. Inside the fuzzy band
the optimum can be reached only by luck. At $64$ bits
$\theta_{\mathrm{ref}}$ lies at $x=-1.23$, between the two levels, so the
constraint at $\beta_{50}$ no longer binds. Its Cholesky factorization fails
at every $b\le60$ and succeeds at $64$ bits through the particular rounding
rather than as a trend. In the one-bit bins around it only about a quarter
to two thirds of the calibration-subset matrices factor, so the optimizer
meets failures on its way in and avoids them only where the rounding
happens to be favorable. Neither optimizer reaches $\theta_{\mathrm{ref}}$:
the Powell fit loses slightly less than the strict prediction, and the best
rejecting-L-BFGS start an order of magnitude less (\cref{tab:optimizers}).
A successful factorization at the optimum therefore does not make a fit
converged.

From $72$ bits $\theta_{\mathrm{ref}}$ is feasible at both levels, yet the
computed objective is still noisy. Its error scales as
$\kappa_2u$~\cref{eq:objerr}; at $\theta_{\mathrm{ref}}$ the measured
$\varepsilon_f$ is about $200$ times below $\kappa_2u$
(\cref{tab:resolution}). A function-value method cannot resolve losses much
below that error~\cref{eq:resolution}, and Powell stops two to three times above
$2\varepsilon_f$ at $72$ and $80$ bits (\cref{tab:sweep}). A gradient method
resolves more finely~\cref{eq:gradresolution}, and the rejecting L-BFGS
reaches a loss three orders of magnitude below Powell's at $72$ bits
(\cref{tab:optimizers}(b), \cref{sec:results:optimizers}).

\Cref{fig:prediction} thus shows three regimes. Well below the threshold, the
constraint at $\beta_{\mathrm{strict}}$ predicts the loss. Within the band,
it gives only an upper bound, and the loss depends on how far the
optimizer gets into the band. Above the band, the loss is set by the
objective error and the optimizer's resolution, which the constraint model
does not describe.

\begin{table}[tbhp]
\centering
\caption{Likelihood loss (nats). (a) Constrained solution predicted at
$h(\theta)\le b+\beta_{\mathrm{strict}}$, best Powell fit of the sweep
(either protocol; continuation at $40$, $53$ and $56$ bits) and SciPy L-BFGS-B over the sweep. (b) Optimizer control:
best (median) over ten shared starts, eight for the gradient methods at
$53$ bits, where two starts fail Cholesky; ``zero'' returns a zero gradient
on a Cholesky failure, ``reject'' rejects the failed trial point.}
\label{tab:optimizers}\label{tab:prediction}
\footnotesize
\setlength{\tabcolsep}{4pt}
\begin{tabular}{l llllll l}
\toprule
$b$ & $40$ & $48$ & $53$ & $56$ & $60$ & $64$ & $72$ \\
\midrule
\multicolumn{8}{l}{(a) prediction and sweep} \\
strict prediction & $17.3$ & $8.90$ & $4.65$ & $2.80$ & $1.11$ & $0.214$ &  $\theta_{\mathrm{ref}}$ feasible \\
best Powell fit & $19.6$ & $16.4$ & $4.72$ & $3.13$ & $1.32$ & $0.202$ & $3.9\times10^{-3}$ \\
L-BFGS-B, zero & $37.9$ & $23.1$ & $23.1$ & $23.1$ & $23.1$ & $8.7$ & $6.0\times10^{-5}$ \\
\midrule
\multicolumn{8}{l}{(b) optimizer control} \\
Powell & & & $4.72$ ($13.8$) & & & $0.202$ ($0.59$) & $3.6\times10^{-3}$ \\
L-BFGS-B, zero & & & $9.47$ ($24.1$) & & & $0.165$ ($9.0$) & $5.9\times10^{-6}$ \\
L-BFGS, reject & & & $3.60$ ($10.9$) & & & $0.0134$ ($0.36$) & $2.5\times10^{-6}$ \\
Newton, reject & & & $3.82$ ($14.3$) & & & $0.316$ ($1.24$) & $1.4\times10^{-2}$ \\
\bottomrule
\end{tabular}
\end{table}

\subsection{Optimizers: failure policy and resolution}\label{sec:results:optimizers}
\begin{figure}[tbhp]
\centering
\includegraphics[width=\textwidth]{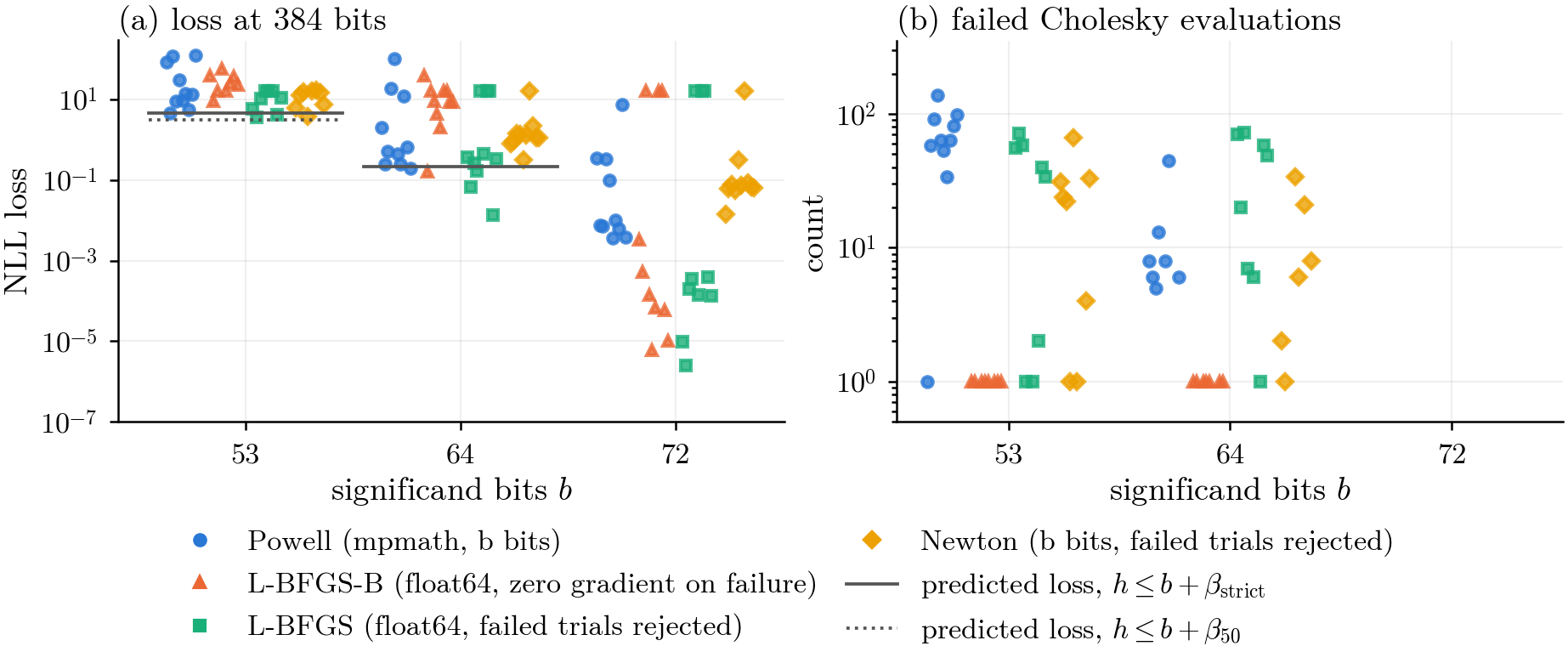}
\caption{Optimizer control. (a) Loss at $384$ bits of each of ten shared
starts for the four methods of \cref{tab:optimizers}(b), omitting the
two starts whose Cholesky factorization fails at $53$ bits, from which the
gradient methods return no fit (Powell, comparing only values, moves off
them); gray: the predicted constrained
loss at $\beta_{\mathrm{strict}}$ (solid) and $\beta_{50}$ (dotted), drawn
only where the level binds. (b) Failed Cholesky evaluations per start on a
log axis: starts without failures are not shown, and none fails at $72$
bits.}
\label{fig:optimizers}
\end{figure}

Below the threshold the failure policy, not the optimizer class, decides
the outcome. scikit-learn's \texttt{GaussianProcessRegressor}~\cite{Pedregosa2011ScikitLearn},
for example, fits hyperparameters with SciPy's
L-BFGS-B~\cite{Byrd1995LimitedMemoryBound,Zhu1997Algorithm778LBFGSB} and
returns a zero gradient at a Cholesky failure, so L-BFGS-B stops at its
first failure; over the sweep it is worse than Powell below $72$ bits and
better from $72$ bits, where no evaluation fails (\cref{tab:optimizers}(a)).
In the optimizer control (protocol in \cref{app:protocols};
\cref{tab:optimizers}(b), \cref{fig:optimizers}) a projected L-BFGS that
instead rejects failed trial points in its backtracking line search gives
the best loss at every precision tested. It also differs from SciPy's
L-BFGS-B in its line search and bound handling, so the control isolates the
failure policy only approximately, but the zero-gradient stop is direct
(\cref{app:values}). At $53$ bits
the median losses of Powell and of both rejecting methods all exceed ten
nats, so the poor independent-protocol fits of \cref{tab:sweep} are a
failure of local optimizers on the penalized objective in general, not of
Powell in particular (caveats in \cref{app:values}). The outlying losses
near $16.5$ nats at $72$ bits (\cref{fig:optimizers}(a)) are not rounding
failures. From three starts (one for Newton) the gradient methods reach a
local optimum with $\sigma_f$ on its upper bound, at $53$ and $64$ bits as
well; the best start, reported in \cref{tab:sweep,tab:optimizers}, avoids
it.

Above the threshold the accuracy is set by how well the optimizer resolves
an objective whose error is proportional to $\kappa_2u$. From $64$ bits,
where the $b$-bit Cholesky factorization first succeeds at
$\theta_{\mathrm{ref}}$, the objective error there follows
$c_f\,\kappa_2 2^{-b}$ with a fitted $c_f$ orders of magnitude below the
worst-case constant of~\cref{eq:objerr}, and the independent-protocol
errors along $v_1$ stay below the function-value bound~\cref{eq:resolution}
except at $96$ bits, where the excess is optimizer noise
(\cref{tab:resolution}). The Newton polish of \cref{sec:setup} removes the
stopping floor of \cref{sec:results:sweep}: the polished fits are at or
below the gradient bound~\cref{eq:gradresolution}, and their hyperparameter
error decreases at roughly $\kappa_2 2^{-b}/\lambda_1$, roughly the square
of the function-value bound $\sqrt{\kappa_22^{-b}/\lambda_1}$. The generality benchmark
therefore measures its thresholds on Newton-polished fits.
\subsection{Generality across kernels, designs and seeds}\label{sec:results:generality}
The generality benchmark (protocol in \cref{app:protocols}) measures, for
$43$ cases, the threshold $b^\ast$: the smallest tested width at which a
criterion holds there and at every higher tested width, for the
independent, Newton-polished fits. A criterion is an accuracy requirement
on the fit, judged against the case's reference; the main one is
$\lVert\delta\theta_{\mathrm{free}}\rVert_\infty\le\tau$, where
$\delta\theta_{\mathrm{free}}$ is $\log\theta_b-\log\theta_{\mathrm{ref}}$ over the
coordinates not on a bound at $\theta_{\mathrm{ref}}$, and $b^\ast(\tau)$ is
its threshold. The level $\tau$ is an accuracy target, not an optimizer
tolerance. The grid starts at $32$ bits, so $b^\ast=32$ means the criterion
held everywhere tested; a criterion still failing at the highest tested
width is \emph{censored}: $b^\ast$ lies above the grid and is not measured.
A \emph{rule} predicts $b^\ast$ from
$\log_2\kappa_2(A_{\theta_{\mathrm{ref}}})$; it is fitted on some cases and
judged by its residuals on the others. The continuation protocol gives similar thresholds at
the loosest tolerances except on the SE Sobol design of seed $3$, discussed
below (\cref{app:values}).

\begin{table}[tbhp]
\centering
\caption{Stage A: $\log_2\kappa_2(A_{\theta_{\mathrm{ref}}})$ (median and
range over seeds $0$--$4$), threshold $b^\ast (\tau)$ for $\lVert\delta\theta_{\mathrm{free}}\rVert_\infty\le\tau$
and the value $\Delta_{53}$ of $\lVert\delta\theta_{\mathrm{free}}\rVert_\infty$ at
$53$ bits (ranges over seeds). In
parentheses: SE Sobol seed $3$, whose $72$- and $80$-bit fits land in a
second mode.}
\label{tab:generality}
\scriptsize
\setlength{\tabcolsep}{3.5pt}
\begin{tabular}{ll c ccc c}
\toprule
kernel & design & $\log_2\kappa_2$ & $b^\ast(10^{-2})$ & $b^\ast(10^{-4})$ & $b^\ast(10^{-6})$ & $\Delta_{53}$ \\
\midrule
SE & clustered & $69.1$ ($66.3$--$70.4$) & $70$--$80$ & $80$--$96$ & $96$ & $0.26$--$2.2$ \\
SE & Sobol & $35.3$ ($29.2$--$39.0$) & $38$--$44$ ($84$) & $44$--$53$ ($84$) & $56$--$72$ ($84$) & $9\times10^{-9}$--$5\times10^{-5}$ \\
Mat\'ern $3/2$ & clustered & $33.7$ ($32.4$--$35.0$) & $32$--$40$ & $40$--$53$ & $53$--$72$ & $1\times10^{-8}$--$8\times10^{-5}$ \\
Mat\'ern $3/2$ & Sobol & $19.7$ ($17.7$--$20.7$) & $32$ & $32$--$48$ & $48$--$53$ & $3\times10^{-12}$--$3\times10^{-7}$ \\
Mat\'ern $5/2$ & clustered & $51.9$ ($48.5$--$52.3$) & $52$--$56$ & $52$--$64$ & $60$--$64$ & $2\times10^{-5}$--$5\times10^{-2}$ \\
Mat\'ern $5/2$ & Sobol & $26.8$ ($25.7$--$27.2$) & $32$ & $32$--$40$ & $40$--$53$ & $3\times10^{-12}$--$4\times10^{-7}$ \\
\bottomrule
\end{tabular}
\end{table}

\begin{figure}[tbhp]
\centering
\includegraphics[width=\textwidth]{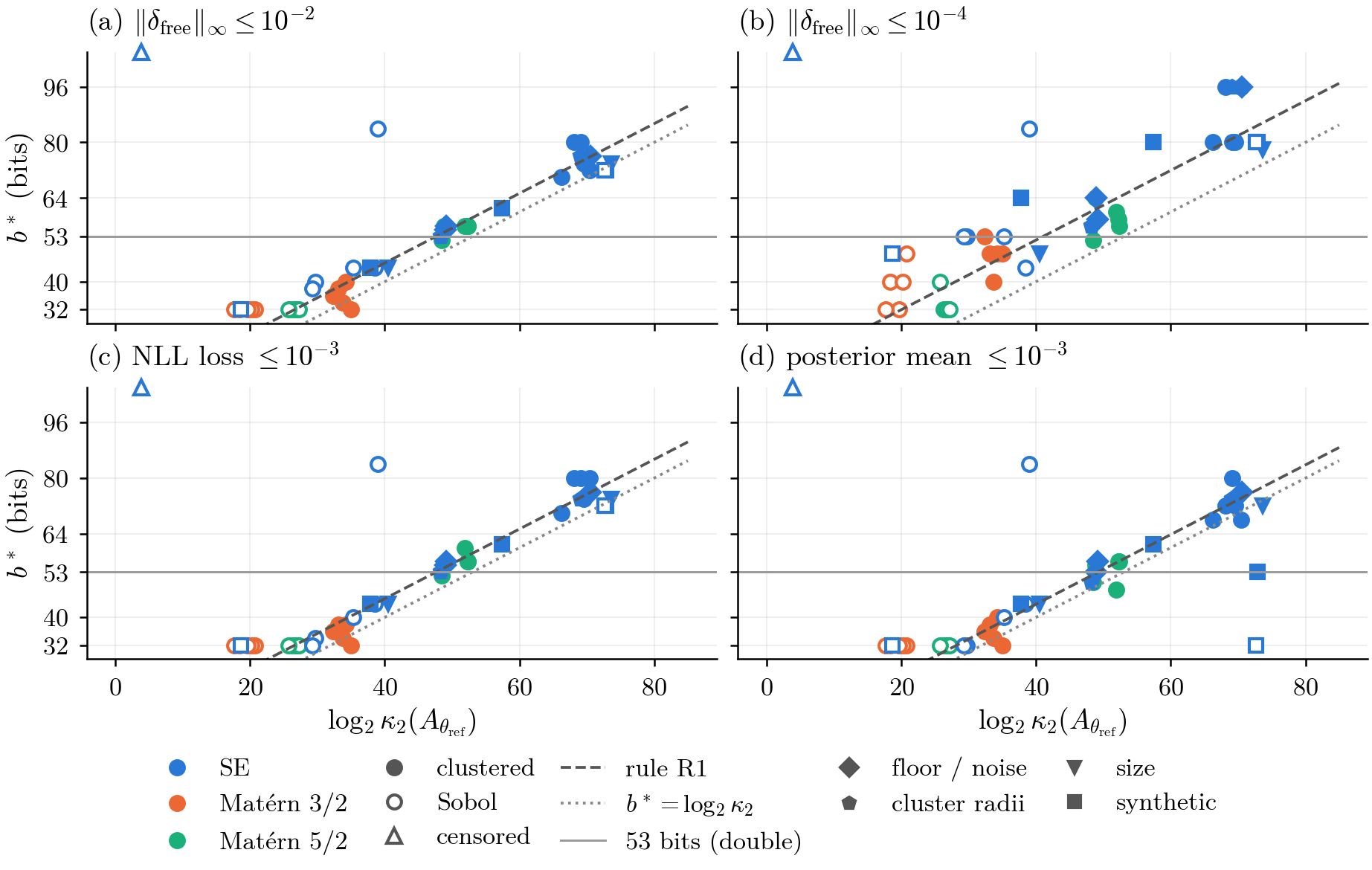}
\caption{Thresholds $b^\ast$ against $\log_2\kappa_2(A_{\theta_{\mathrm{ref}}})$
for (a, b) $\lVert\delta\theta_{\mathrm{free}}\rVert_\infty\le10^{-2}$ and
$10^{-4}$, (c) likelihood loss $\le10^{-3}$ nats and (d) posterior-mean change
$\le10^{-3}$ (relative $L^2$). Color: kernel; filled: clustered, open:
Sobol design; marker shape: Stage A or Stage B block; triangles on the top
edge: censored ($b^\ast>256$). The open SE circle at $84$ bits and the
censored case are the two multimodal cases.}
\label{fig:p2-threshold}
\end{figure}

\paragraph{Thresholds track the conditioning at the reference}
Across the benchmark the threshold for
$\lVert\delta\theta_{\mathrm{free}}\rVert_\infty\le10^{-2}$ follows
$\log_2\kappa_2(A_{\theta_{\mathrm{ref}}})$
(\cref{fig:p2-threshold,tab:generality}): kernel and design enter through
this conditioning, which is worst for the SE kernel on the clustered designs
and best for the Mat\'ern $3/2$ kernel, whose $53$-bit fits meet
$\tau=10^{-2}$ in all ten of its cases but $\tau=10^{-6}$ only in some. Of
the $43$ cases $19$ have $b^\ast(10^{-2})>53$, $17$ because $53$ bits are
insufficient and two through mode selection (below). The other $24$, which
double precision already resolves, serve as negative controls, on which a
rule must not call for extended precision. On the SE clustered designs the
reference covariance does not factor at $53$ bits, so the implicit
constraint binds and the polished $53$-bit fits lose several nats
(\cref{app:values}); on the Mat\'ern $5/2$ clustered designs it factors,
but objective and gradient errors of the size estimated by~\cref{eq:objerr}
limit how closely even the polished $53$-bit fits approach the reference.

\paragraph{Rules}
Rule R1 is $b^\ast=\log_2\kappa_2(A_{\theta_{\mathrm{ref}}})+c$, with $c$ the
median offset over Stage A seeds $0$--$2$, assessed on the $25$ held-out
cases (seeds $3$--$4$ and Stage B; \cref{tab:rules}). The offset grows by
about the change in $\log_2(1/\tau)$, $\log_2100\approx6.6$ bits, from
$10^{-2}$ to $10^{-4}$, and by about twice that from $10^{-4}$ to
$10^{-6}$. Rule R2, which adds the curvature term
$\log_2(1/(\lambda_1\tau^2))$ of \cref{sec:theory:resolution} with a
constant fitted per criterion, does no better.

\begin{table}[tbhp]
\centering
\caption{Rules R1 and R2: offset $c$ of R1 (bits) and mean absolute
residual (MAR, bits) on the held-out cases, $24$ of the $25$ because the
censored $d=4$ Sobol case has no residual. ``Excl.'' also omits SE Sobol
seed $3$; ``max'' is the largest absolute residual.}
\label{tab:rules}
\footnotesize
\setlength{\tabcolsep}{4pt}
\begin{tabular}{l r rrrr}
\toprule
 & & \multicolumn{3}{c}{R1} & R2 \\
\cmidrule(lr){3-5}
criterion & $c$ & MAR & MAR, excl. & max, excl. & MAR \\
\midrule
$\lVert\delta\theta_{\mathrm{free}}\rVert_\infty\le10^{-2}$ & $5.4$ & $4.2$ & $2.7$ & $9$ & $4.5$ \\
$\lVert\delta\theta_{\mathrm{free}}\rVert_\infty\le10^{-4}$ & $12.0$ & $7.1$ & $6.0$ & $17$ & $7.1$ \\
$\lVert\delta\theta_{\mathrm{free}}\rVert_\infty\le10^{-6}$ & $26.2$ & $7.9$ & $7.4$ & $22$ & $7.7$ \\
\bottomrule
\end{tabular}
\end{table}

\paragraph{Counterexamples}
Three held-out cases are counterexamples to R1 at $10^{-2}$ (criteria in
\cref{app:protocols}, values in \cref{app:values}). One is a Mat\'ern $3/2$ Sobol design whose predicted
threshold lies below the grid floor. The other two are set by mode
selection, not arithmetic: on the SE Sobol design of seed $3$ the fits at
$72$ and $80$ bits end in a second mode although those at $53$--$64$ bits and
from $84$ bits reach the reference, and on the best-conditioned case, a
synthetic $d=4$ Sobol design ($\log_2\kappa_2\approx4$), every fit from $32$
to $256$ bits ends in a non-reference mode, so every criterion on
the fitted parameters, loss and posterior is censored. This case is an
outlier not because it needs more than $256$ bits, but because no precision
changes the mode reached from the shared starts: at $\theta_{\mathrm{ref}}$
Cholesky succeeds and the objective error is below $10^{-6}$ nats at every
tested width. A threshold measured from two fixed starts on a
multimodal likelihood therefore need not be monotone in $b$, and its excess
over the rule is not arithmetic insufficiency.

\paragraph{Geometry, size, noise and dimension}
Stage B varies one ingredient of the seed-$3$ SE problem at a time
(\cref{tab:stageb}). The noise floor caps the conditioning, so pulling the
cluster in leaves the threshold near that of the seed-$3$ SE problem itself,
while spreading it, raising the floor or adding observation noise lowers
conditioning and threshold alike. Lowering the floor to $10^{-14}$ moves
the fitted parameters more than raising the precision beyond $72$ bits
does (\cref{app:values}), so sensitivity to the floor deserves reporting
alongside sensitivity to precision. On
the one-dimensional synthetic designs, whose kernel matrices $K$ are the
worst conditioned, the posterior mean is within $10^{-3}$ of the reference
at $53$ bits although the hyperparameters are not. Each setting is a single
case, consistent with R1 but not evidence of a trend in $n$, $d$ or noise.

\subsection{Consequences: posterior and closed-loop active learning}\label{sec:results:posterior}\label{sec:results:bo}
In the test problem the precision-induced hyperparameter changes barely move
the posterior mean near the BO cluster, where data pin it down, but with
double-precision hyperparameters the posterior standard deviation there
averages $8.4$ times the reference's near the cluster (\cref{app:values}). This is a
difference from the reference posterior, not a calibration claim.

In closed loop the surrogate also chooses the design. Each active-learning
run adds $60$ points to $8$ Sobol points, refitting $\theta$ by maximum
likelihood before every selection. Each selection follows the active
learning Cohn (ALC) criterion~\cite{Cohn1996NeuralNetworkExploration,Seo2000GPActiveDataSelection}: it
adds the candidate that most reduces the posterior variance integrated over
the domain, the integrated mean squared error (IMSE) of the predictor
\cite{Sacks1989DesignAnalysisComputerExperiments}. Runs use $b=53$ and $64$
bits against a $200$-bit reference on three seeds with shared candidate sets (\cref{app:protocols}), so runs on the same
data differ only through their arithmetic. Performance is the relative $L^2$
error of the posterior mean, a measure of recovery and not of optimization
regret. Besides Branin, two objectives pull the MLE toward the flat limit
of \cref{sec:background}: a rotated anisotropic quadratic bowl and the
Forrester function of the first input alone (``inactive dimension'').

\begin{figure}[tbhp]
\centering
\includegraphics[width=\textwidth]{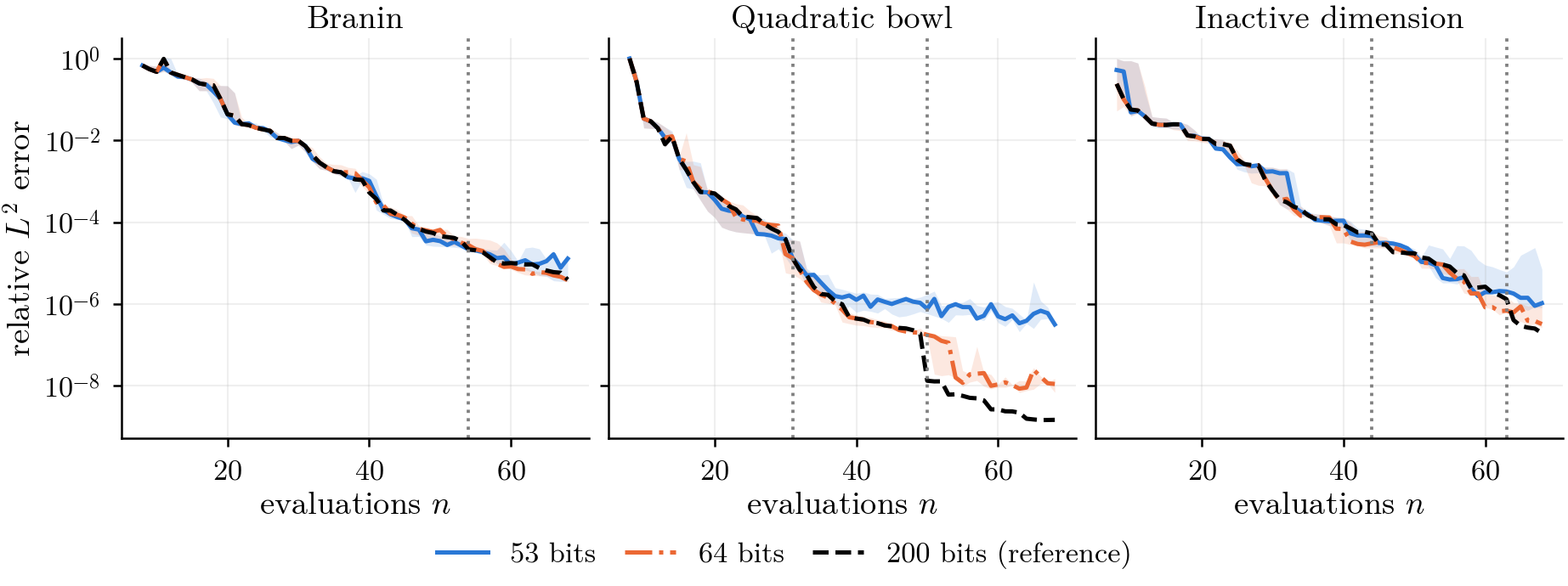}
\caption{Relative $L^2$ error of the posterior mean against the number of
evaluations (median and range over seeds $0$--$2$). Vertical dotted lines:
the reference's median $\kappa_2(A_\theta)$ first reaches $2^{53}$ and
$2^{64}$ (the latter not reached on Branin).}
\label{fig:bo-l2}
\end{figure}

\paragraph{Conditioning}
On these two objectives the reference conditioning passes $2^{53}$ about
halfway and $2^{64}$ before the end of the run (dotted lines in
\cref{fig:bo-l2}); Branin barely reaches the regime where $53$ bits bind. The
$b$-bit fits instead stop a few bits above their working precision by
fitting a larger noise, the implicit nugget of \cref{sec:results:regimes},
so the conditioning at a low-precision fit understates the reference's.

\paragraph{Recovery error}
The errors agree until the reference conditioning passes the cap of the
working precision and separate after it (\cref{fig:bo-l2}): on the quadratic
the $53$-bit error stalls and ends a factor of $214$ above the reference.
The gaps are smaller on the inactive objective, where crossing $2^{53}$ is
not enough to separate the runs, and on Branin, where the $64$-bit run
matches the reference.

\paragraph{Hyperparameters versus design}
\Cref{tab:bo-decomp} recomputes the final error at $200$ bits with one
ingredient of the $b$-bit run changed at a time. Evaluating the $b$-bit model
itself at $200$ bits barely changes the error (\cref{app:values}), so the
final solve is not the cause. On the quadratic, $\theta_b$ on the
reference design stays well above the reference error, whereas a $200$-bit
refit on the $b$-bit design nearly recovers it; the excess error thus
travels with the
hyperparameters, through the implicit nugget, and not with the design, which
leaves the reference's early but remains space filling. Branin at $53$ bits
and the inactive objective on two of three seeds show the same pattern more
mildly; on inactive seed $0$ the design $X_{53}$ is several times less
informative than the reference design at either $\theta$, and the refit on
it is poor although it is the best mode found (\cref{sec:results:remedies}).

\begin{table}[tbhp]
\centering
\caption{Final relative $L^2$ error ($n=68$), median over seeds $0$--$2$, with
one ingredient of the $b$-bit run changed at a time. The first row is the
$b$-bit run itself; the other rows are evaluated at $200$ bits, the last being
the reference run. ``Refit'' is an MLE at $200$ bits warm-started at
$\theta_b$.}
\label{tab:bo-decomp}
\small\setlength{\tabcolsep}{4pt}
\begin{tabular}{lccccc}
\toprule
& \multicolumn{2}{c}{quadratic} & \multicolumn{2}{c}{inactive} & Branin \\
design, $\theta$ & $b=53$ & $b=64$ & $b=53$ & $b=64$ & $b=53$ \\
\midrule
$X_b$, $\theta_b$ ($b$-bit run) & $3.2\times10^{-7}$ & $1.1\times10^{-8}$ & $1.05\times10^{-6}$ & $3.3\times10^{-7}$ & $1.3\times10^{-5}$ \\
$X_{\mathrm{ref}}$, $\theta_b$ & $1.3\times10^{-7}$ & $6.3\times10^{-9}$ & $1.8\times10^{-6}$ & $3.0\times10^{-7}$ & $8.3\times10^{-6}$ \\
$X_b$, refit $\theta$ & $2.8\times10^{-9}$ & $1.7\times10^{-9}$ & $2.2\times10^{-7}$ & $3.2\times10^{-7}$ & $4.7\times10^{-6}$ \\
$X_{\mathrm{ref}}$, $\theta_{\mathrm{ref}}$ & $1.5\times10^{-9}$ & $1.5\times10^{-9}$ & $2.0\times10^{-7}$ & $2.0\times10^{-7}$ & $4.0\times10^{-6}$ \\
\bottomrule
\end{tabular}
\end{table}

\subsection{Remedies on fixed designs}\label{sec:results:remedies}
\Cref{tab:remedies} compares the remedy arms (protocol in \cref{app:protocols}) on
stored data, not inside the learning loop; counts, timings and exceptions are
in \cref{app:values}. The $53$-bit arm is a fresh fit from the two shared
starts, not the in-loop fit of \cref{tab:bo-decomp}.

\emph{Extended precision} works from $80$ significand bits, not from the
$64$ of x87 extended precision: with the whole
likelihood, kernel entries included, at $80$ bits the recovery error reaches
the reference arm's on every design, with no Cholesky failure or negative
predictive variance, whereas the $64$-bit arm has both and still loses
likelihood. Raising the precision of the Cholesky alone is not enough:
at $\theta_{\mathrm{ref}}$, $A$ with float64 entries is already indefinite,
so mixed-precision iterative
refinement~\cite{Higham2022MixedPrecision,Carson2018IterativeRefinement},
which refines the solve but keeps the float64 entries, cannot recover it.

\emph{Jitter} restores factorability, but on the quadratic and inactive
designs, where the reference errors are tiny, the jittered posterior recovers
the function worse than the no-jitter one at every $j$, monotonically in
$j$; on Branin, where the no-jitter $53$-bit fit is poor, a small jitter improves
on it. The nugget bias
of~\cite{Ranjan2011ComputationallyStable,Andrianakis2012EffectNugget} is thus
large relative to the reference errors, although the absolute recovery
errors on those designs ($10^{-9}$ to a few $10^{-4}$) may or may not
matter in an application.

A $384$-bit \emph{refit} warm-started at the stored $53$-bit fit costs one
high-precision fit instead of one per acquisition step, and comes within
$10\%$ of the recovery error of the best $384$-bit fit found on $15$ of the
$18$ designs ($X_{53}$ and $X_{64}$, three objectives, three seeds) as
fitted and on $17$ after a Newton polish; on the
remaining Branin design it lands in a wrong mode, so a second start is
needed. On inactive seed $0$ this $384$-bit refit has the lowest
NLL found by fifteen $384$-bit starts on $X_{53}$ (the best found, not a
certified global optimum), yet it recovers the function several times worse
than a second mode a few nats above it: a better likelihood does not
guarantee better recovery. scikit-learn's \texttt{GaussianProcessRegressor}
($1.7.2$)~\cite{Pedregosa2011ScikitLearn}, matched to the arms' kernel,
bounds and starts, reproduces their failure modes: with $\alpha=0$ its
L-BFGS-B stops at the first Cholesky failure
(\cref{sec:results:optimizers}), and with $\alpha=10^{-10}$ its recovery
error matches that of the $53$-bit arm with $j=10^{-10}$ on three of the
four designs, although its likelihood differs. Its documented default, a
fixed kernel with unit parameters, is not a fit at all.

\begin{table}[tbhp]
\centering
\caption{Remedy arms (float64 Powell outer optimizer, two shared starts).
Loss: likelihood loss of the intended (no-jitter) NLL at $384$ bits
(nats); rec.: relative $L^2$ error of the arm's own posterior mean. The $n=68$
columns are medians over seeds $0$--$2$. The ``$384$ bits, shared starts''
row is the reference arm; its loss on the inactive objective is explained in
\cref{app:values}.}
\label{tab:remedies}
\scriptsize
\setlength{\tabcolsep}{2.5pt}
\begin{tabular}{l rr rr rr r}
\toprule
& \multicolumn{2}{c}{$n=30$ Branin} & \multicolumn{2}{c}{quadratic, $X_{53}$} & \multicolumn{2}{c}{inactive, $X_{53}$} & Branin, $X_{53}$ \\
\cmidrule(lr){2-3}\cmidrule(lr){4-5}\cmidrule(lr){6-7}\cmidrule(lr){8-8}
arm & loss & rec. & loss & rec. & loss & rec. & rec. \\
\midrule
$53$ bits, $j=0$ & $13.3$ & $0.147$ & $166$ & $4.3\times10^{-7}$ & $32$ & $1.05\times10^{-6}$ & $2.4\times10^{-5}$ \\
$53$, $j=10^{-12}$ & $8.47$ & $0.109$ & $72$ & $5.0\times10^{-7}$ & $20$ & $2.1\times10^{-6}$ & $4.7\times10^{-6}$ \\
$53$, $j=10^{-10}$ & $5.57$ & $0.0431$ & $61$ & $3.3\times10^{-6}$ & $35$ & $1.1\times10^{-5}$ & $1.5\times10^{-5}$ \\
$53$, $j=10^{-8}$ & $2.18$ & $0.0507$ & $127$ & $4.7\times10^{-5}$ & $55$ & $5.3\times10^{-5}$ & $5.0\times10^{-5}$ \\
$53$, $j=10^{-6}$ & $1.87$ & $0.0613$ & $65$ & $1.4\times10^{-4}$ & $21$ & $2.8\times10^{-4}$ & $7.7\times10^{-4}$ \\
$64$ bits & $0.48$ & $0.0281$ & $37$ & $3.4\times10^{-8}$ & $9.0$ & $3.3\times10^{-7}$ & $4.7\times10^{-6}$ \\
$80$ bits & $2.0\times10^{-4}$ & $0.0273$ & $0.34$ & $2.75\times10^{-9}$ & $4.5$ & $1.51\times10^{-7}$ & $4.71\times10^{-6}$ \\
$384$ bits, shared starts & $5.2\times10^{-11}$ & $0.0273$ & $0.23$ & $2.73\times10^{-9}$ & $4.5$ & $1.51\times10^{-7}$ & $4.71\times10^{-6}$ \\
$384$ bits, refit from $\theta_{53}$ & $0.082$ & $0.0280$ & $0.22$ & $2.78\times10^{-9}$ & $1.6\times10^{-3}$ & $8.2\times10^{-7}$ & $4.72\times10^{-6}$ \\
\bottomrule
\end{tabular}
\end{table}

\paragraph{Choosing a remedy}\label{sec:results:choosing}
The precision margins apply to the conditioning at the reference optimum:
no sampled or reference covariance with $\lambda_{\min}(R_\theta)\ge2.3u$,
and no sampled covariance with $\kappa_2<2^{b+3}$, failed to factor here
(the second edge is less sharp because $\lambda_{\max}(R_\theta)$ varies;
\cref{app:values}), and an accurate fit needs the further offset of rule R1
(\cref{tab:rules}) unless the likelihood is multimodal. At a low-precision
fit $\log_2\kappa_2(A_{\theta_b})$ already reflects the implicit constraint,
sitting a few bits above $b$, so it is a warning sign that the constraint
binds rather than a validated estimate of the precision needed. When double
precision falls short, the comparisons favor promoting the whole
likelihood, kernel entries included, and rejecting failed trial points
rather than returning a zero gradient (\cref{sec:results:optimizers}). A
nugget is the cheap alternative, desirable on statistical grounds according
to Gramacy and Lee~\cite{Gramacy2012CasesNugget}, but for a deterministic
simulator its cost in recovery error where the reference interpolates
closely must be weighed against the application. Stable bases
(RBF-QR~\cite{Fornberg2011StableGaussianRBF} and the Hilbert--Schmidt
SVD~\cite{Fasshauer2012StableGaussianRBF}, applied to the likelihood
in~\cite{McCourt2017StableLikelihoodGRF}) and reparameterizations or priors
that avoid degenerate correlation matrices~\cite{Gu2018RobustGaSP} are not
compared: the former are kernel specific, unlike promoting the precision,
and the latter change the estimator rather than its arithmetic.

\section{Limitations and conclusions}\label{sec:discussion}

The sweep and the mechanism experiments use one design, one seed, one
kernel and fixed optimizer starts; outside the test problem the benchmark
checks only the safe side of the factorability boundary. The sharp constraint
of~\cref{eq:constrained} describes the fits only well below the threshold,
since near it the factorable set has a fuzzy edge, and \cref{prop:kkt}
covers only the fixed-noise branch at small violations, which the free
branch follows until $\sigma_n$ leaves its floor. The
active-learning runs use three seeds, one acquisition criterion and a
float64 outer optimizer, and on some objectives the box constraints set the
reference conditioning. The remedies are compared on fixed designs, not
inside the loop, and Bayesian-optimization regret is not studied. The
arithmetic is an mpmath emulation of significand width only, without
exponent range, subnormals or fused operations. Its run times, reported
only for the remedy refits (\cref{app:values}), do not measure the speed of
extended precision in hardware or optimized libraries, so the paper makes no
claims about the computational cost of extended precision.

\paragraph{Conclusions}\label{sec:conclusion}
In a routine noiseless kriging problem with a BO-style design, IEEE double
precision does not deliver the maximum-likelihood hyperparameters, and the
fits converge only once the significand width exceeds the base-2 logarithm
of the covariance condition number at the optimum. Cholesky failure acts as
an implicit conditioning constraint, caused mostly by rounding of the
kernel entries and amplified along weakly identified directions; because
the constrained solution predicted from the working precision alone is
close to the best fits at most precisions below the threshold, the poorest
double-precision fits are optimizer failures on a penalized objective, and
the handling of failed factorizations matters more than the optimizer
class. In the benchmark the bits needed follow the conditioning at the
reference optimum plus a tolerance-dependent offset; at the loosest
tolerance the exceptions are the grid floor and mode switches of the
optimizer, not arithmetic. The consequences are largest for uncertainty:
double-precision hyperparameters inflate the posterior standard deviation
near the design cluster, and in closed-loop active learning on a
flat-limit objective the implicit nugget can stall recovery once the
design's conditioning outgrows the working precision. On fixed designs,
promoting the whole likelihood, kernel entries included, to a significand
wider than that of x87 extended precision recovers the reference accuracy,
whereas a fixed jitter biases the posterior where the reference
interpolates closely and a single high-precision refit, though it replaces
one fit per step, is not always reliable. Open are a stochastic model of the boundary band, a theory
for the branch after the noise bound releases, remedies run inside the
learning loop, and regret in Bayesian optimization.

\appendix

\section{Proof of \texorpdfstring{\cref{prop:kkt}}{Proposition}}\label{app:kkt}

\begin{proof}
Write $f=\mathrm{NLL}$ and $c=b+\beta_g=g(\theta^\ast)-r_b$ for the level;
$q>0$ because
$H\succ0$ and $s\ne0$.

\emph{Step 1: a smooth branch of KKT points.} Zeros with $\mu\ge0$ of
$F(\theta,\mu,r)=\bigl(\nabla f(\theta)+\mu\nabla g(\theta),\
g(\theta)-g(\theta^\ast)+r\bigr)$ are the KKT points
of~\cref{eq:constrained} with the constraint active and multiplier $\mu$.
The map $F$ is $C^2$, $F(\theta^\ast,0,0)=0$, and its Jacobian with respect
to $(\theta,\mu)$ there is
$J=\bigl[\begin{smallmatrix}H&s\\s^\top&0\end{smallmatrix}\bigr]$, with
$\det J=-q\det H\ne0$ (Schur complement). By the implicit function theorem
there are $r_1>0$ and $C^2$ functions $\theta(r)$, $\mu(r)$ on $(-r_1,r_1)$
with $\theta(0)=\theta^\ast$, $\mu(0)=0$ and $F(\theta(r),\mu(r),r)=0$.
Differentiating at $r=0$ gives $H\theta'(0)+\mu'(0)s=0$ and
$s^\top\theta'(0)=-1$, so $\mu'(0)=1/q$, $\theta'(0)=-H^{-1}s/q$ and, by
Taylor's theorem, $\theta(r)=\theta^\ast-rH^{-1}s/q+O(r^2)$ and
$\mu(r)=r/q+O(r^2)$. The violation $r_b$ of \cref{prop:kkt} is the value
of $r$ that matters; $r$ is kept free here only to differentiate.

\emph{Step 2: the branch gives the local minimizer.} Let
$\lambda_H=\lambda_{\min}(H)$. Choose $\rho>0$ such that the closed ball
$\bar B$ of radius $\rho$ about $\theta^\ast$ lies in $U$ and, on $\bar B$,
$\lVert\nabla^2f-H\rVert_2\le\lambda_H/4$, and let
$M=\max_{\bar B}\lVert\nabla^2g\rVert_2$, which is finite because $\bar B$ is
compact; set $\mu_0=\lambda_H/(4M)$ ($\infty$ if
$M=0$). By Weyl's inequality the Lagrangian
$\mathcal{L}(\theta,\mu)=f(\theta)+\mu\,(g(\theta)-c)$ has
$\nabla^2_\theta\mathcal{L}\succeq(\lambda_H/2)I$ for $\theta\in\bar B$ and
$0\le\mu\le\mu_0$. By Step 1 there is $r_0\in(0,r_1]$ such that, for
$0<r_b<r_0$, $\theta(r_b)$ lies in the open ball $N$ of radius $\rho$ about
$\theta^\ast$ and $\mu=\mu(r_b)\in(0,\mu_0]$. Then $\mathcal{L}(\cdot,\mu)$ is
strictly convex on $\bar B$ and stationary at $\theta(r_b)$, hence uniquely
minimized there. For every feasible $\theta\in N$, $\mu>0$ and
$g(\theta)\le c=g(\theta(r_b))$ give
$f(\theta)\ge\mathcal{L}(\theta,\mu)\ge\mathcal{L}(\theta(r_b),\mu)=f(\theta(r_b))$,
with equality only if $\theta=\theta(r_b)$. So $\theta^{\mathrm{c}}_b=\theta(r_b)$ is the
unique minimizer of~\cref{eq:constrained} over $N$, the constraint is active
there, and the first expansion in~\cref{eq:kktloss} is that of Step 1.

\emph{Step 3: the likelihood loss.} Let $\delta\theta=\theta^{\mathrm{c}}_b-\theta^\ast=O(r_b)$.
Since $\nabla f(\theta^\ast)=0$ and $f$ is $C^3$ on $\bar B$,
$f(\theta^{\mathrm{c}}_b)-f(\theta^\ast)=\tfrac12\delta\theta^\top H\,\delta\theta+O(r_b^3)$, and
substituting $\delta\theta=-r_bH^{-1}s/q+O(r_b^2)$ gives
$\tfrac12\delta\theta^\top H\,\delta\theta=r_b^2s^\top H^{-1}s/(2q^2)+O(r_b^3)=r_b^2/(2q)+O(r_b^3)$.
\end{proof}

\paragraph{Smoothness of $g(\theta)$}
Since $A_\theta$ is real analytic in the log-parameters, a simple eigenvalue
$\lambda$ of it, with unit eigenvector $v$, is analytic too (implicit
function theorem applied to $(A_\theta-\lambda I)v=0$, $v^\top v=1$). Hence
$g(\theta)=\log_2\lambda_{\max}(A_\theta)-\log_2\lambda_{\min}(A_\theta)$ is $C^3$
in $\theta$ near $\theta^\ast$ when the extreme eigenvalues of $A_{\theta^\ast}$ are
simple; at a multiple extreme eigenvalue $g(\theta)$ is in general not
differentiable and the proposition does not apply as stated.

\section{Protocols}\label{app:protocols}

\paragraph{Design and sweep}
The $10$ clustered points lie at radii shrinking geometrically from
$5\times10^{-2}$ to $10^{-3}$ around a minimizer. The sweep tests $b=11$, $16$, $20$, $24$, $32$, $40$, $48$, $53$, $56$,
$60$, $64$, $72$, $80$, $96$, $128$, $160$, $200$ and $256$; at $11$ bits all $72$ evaluations failed Cholesky, and \cref{tab:sweep} omits
$11$ and $20$ bits (the $20$-bit fit loses $53.1$ nats). A failure returns
the penalty $10^{30}$. The optimizer is a line-by-line port of SciPy's
bounded Powell method~\cite{Powell1964EfficientMinimization} to mpmath, so comparisons, line searches and iterates
are never rounded to float64; SciPy's absolute tolerance floors are scaled
by $2^{53-b}$, so that at $b=53$ the port reproduces SciPy exactly. It uses
$\mathrm{xtol}=10^{-8}$, $\mathrm{ftol}=10^{-12}$, at most $100$ iterations
and the starts $(1,0.3,0.3,10^{-2})$ and $(1,1,1,10^{-4})$ in
$(\sigma_f,\ell_1,\ell_2,\sigma_n)$.

\paragraph{Newton polish}
The projected Newton method uses the analytic gradient and Hessian
of~\cref{eq:nll} in the log-parameters, formed from the same $b$-bit
Cholesky factor as the NLL and checked against finite differences. A
parameter at a bound whose gradient points outward (here $\sigma_n$) is
held there; backtracking steps are accepted if they lower the computed NLL
or, at its noise level, the free gradient. A $b$-bit Powell fit is
polished only when Powell's stopping test, not arithmetic, ended the
search:
$\lvert\mathrm{NLL}_b(\theta_b)-\mathrm{NLL}_{\mathrm{ref}}(\theta_b)\rvert\le\mathrm{ftol}\cdot\lvert\mathrm{NLL}_b(\theta_b)\rvert$.
The polished reference has a free gradient of $10^{-95}$. Metric (iii)
evaluates both posteriors at $384$ bits.

\paragraph{Factorability map and branches}
The $686$ points are the sweep's fits, the constrained solutions, and rays
from $\theta_{\mathrm{ref}}$ along the three Hessian eigenvectors and $24$
seeded random directions in the free log-parameters, on both sides of the
boundary. At each $b\in\{40,48,53,56,60,64,68,72,80\}$ the map records
whether the $b$-bit pipeline succeeds, the failing pivot and the NLL error.
The splits form entries at $b$ bits and factor at $384$; form at $384$
bits, round once to $b$ and factor at $b$; or factor those rounded entries
at $384$. As a control, entries formed and factored at $384$ bits factor
at every point (all $6174$ trials), so each $b$-bit failure is due to
precision and not to an indefinite matrix. The band is calibrated on the Hessian rays and the even-numbered
random rays ($2250$ trials) and checked on the rest. The $95\%$ edge is the
lowest one-bit bin edge in $x$ above which every bin of the calibration
subset has a success fraction of at least $0.95$. Constrained problems
are solved by SciPy's SLSQP from $\theta_{\mathrm{ref}}$, the shared starts, eight
seeded starts, neighboring levels and known feasible fits, then polished
to KKT points at $384$ bits (residuals below $3\times10^{-87}$, positive
definite reduced Hessian) and rechecked at $512$ bits.

\paragraph{Optimizer control}
At $53$, $64$ and $72$ bits, ten shared starts are run with the mpmath
Powell method; SciPy's float64 L-BFGS-B with a zero gradient on a Cholesky
failure, as in scikit-learn's default fit; a projected L-BFGS with Armijo
backtracking that rejects failed trial points, clips to the box and freezes
coordinates at a bound; and a $b$-bit Newton method with the same
rejection. The L-BFGS-B sweep row of \cref{tab:optimizers}(a) uses
SciPy's defaults with the NLL and gradient at $b$ bits.

\paragraph{Generality benchmark}
The $43$ cases use $21$ data sets. Stage A fits the three kernels to
Branin with $n=30$ on clustered designs built as in \cref{sec:setup} and on
$30$ Sobol points, seeds $0$--$4$; the seed-$3$ SE clustered case is the
sweep problem. Stage B (\cref{tab:stageb}) uses cluster fraction $1/3$ for
$n=20$ and $50$, observation noise of standard deviation $10^{-6}$ at floor
$10^{-9}$, and the synthetic objective
$f(w)=\sum_j\sin(2\pi w_j/p_j)+\tfrac12\prod_j\cos(\pi w_j/p_j)$ with
periods $p=(0.5,1,1.5,2)$ truncated to $d$ inputs. Each reference is the
lowest Newton-polished $384$-bit Powell fit from the two shared starts and
eight log-uniform starts, checked at $512$ bits. Fits use the sweep's
widths from $32$ to $160$ bits (coarser for the $n$ and synthetic blocks),
refined at two- or four-bit spacing near the first threshold and extended
to $256$ bits when needed; each is a $b$-bit Powell fit followed by a
$b$-bit Newton polish, applied without the stopping-test condition used in
the sweep. The
criteria are the error over the reference's free coordinates, with the same
active set ($10^{-2}$, $10^{-4}$, $10^{-6}$), the likelihood loss
($10^{-3}$, $10^{-6}$ nats), the relative $L^2$ change of the posterior mean
and standard deviation ($10^{-3}$, $10^{-5}$), the objective error at
$\theta_{\mathrm{ref}}$ ($10^{-3}$, $10^{-6}$, $10^{-9}$ nats) and Cholesky
success at $\theta_{\mathrm{ref}}$. A held-out case is a counterexample when
its residual exceeds $8$ bits, when it is censored but predicted inside the
grid, or when the rule's double-precision verdict is wrong.

\paragraph{Active learning}
Each step adds, among $1024$ Sobol candidates that depend only on the seed
and the iteration, the point maximizing
$\sum_{z\in Z}\mathrm{cov}_n(z,x)^2/(\mathrm{var}_n(x)+\sigma_n^2)$ over
$256$ fixed Sobol points $Z$. Likelihood, acquisition and the error
$\lVert m_n-f\rVert_2/\lVert f-\bar f\rVert_2$ ($41\times41$ grid) are
evaluated at $b$ bits, but the outer optimizer is SciPy's float64 Powell
(at most $20$ iterations, $\mathrm{xtol}=10^{-5}$, $\mathrm{ftol}=10^{-8}$,
from the shared starts and the previous fit). Rerun at $384$ bits, this
protocol stops $2.4$--$5.8$ nats above the polished optimum on the inactive
designs (the $5.8$ on seed $0$, in a second mode) and up to $1.3$ nats on the quadratic. In the reference fits
Branin's $\sigma_f$, the quadratic's $\sigma_f$ and $\sigma_n$, and the
inactive objective's $\ell_2$ (and $\sigma_n$ on two seeds) sit on their
bounds. The quadratic is
$f(w)=30\,(Q(w-w_0))^\top\mathrm{diag}(1,10)\,(Q(w-w_0))$ with
$w_0=(0.37,0.61)$ and $Q$ a $30^\circ$ rotation; the Forrester
function~\cite{Forrester2008SurrogateModelling} is
$f(w)=(6w_1-2)^2\sin(12w_1-4)$.

\paragraph{Remedies}
The $n=68$ datasets are $X_{53}$ and $X_{64}$ for the three objectives and
seeds $0$--$2$. The jitter $j\in\{10^{-12},10^{-10},10^{-8},10^{-6}\}$ is a
fixed variance added to the training covariance, $A=K+(\sigma_n^2+j)I$,
not at test points, not scaled by $\sigma_f^2$ and not part of the
statistical model. Extended precisions are $64$, $80$, $106$ and $113$ bits, the last two
the significands of double-double~\cite{Dekker1971FloatingPointTechnique}
and IEEE binary128~\cite{IEEE2019Std754} (omitted from \cref{tab:remedies}, recover as the reference
arm does). Arms use the sweep's tolerances for $n=30$ and the closed-loop
ones for $n=68$. The loss uses the intended NLL (no jitter, $384$ bits);
the recovery error uses the arm's own posterior (its bits, its $j$).
Fifteen $384$-bit fits on inactive seed $0$, $X_{53}$, start from the
stored fits, the shared starts, eight seeded starts and the warm refit.
scikit-learn's \texttt{GaussianProcessRegressor} is run in its default and
matched to the arms with training jitter $\alpha=0$ and $10^{-10}$.

\section{Supporting values}\label{app:values}

\paragraph{Local quantities}
At $\theta_{\mathrm{ref}}$ (free log-parameters, $384$ bits) $H$ has
eigenvalues $2.59$, $192$, $1538$ (condition number about $593$),
$v_1\approx(0.885,0.091,0.456)$, $s=(0.744,13.56,3.90)$ and $q=5.36$. At
the floor the slopes in $\log\sigma_n$ are $0.129$ for the NLL and $-0.744$
for $g(\theta)$, so $\mu\approx r_b/q$ exceeds their ratio for $r_b\gtrsim0.93$; the
computed multiplier of the $\sigma_n$ bound changes sign earlier, near
$r_b\approx0.55$ (\cref{tab:branch}), because
$\partial\mathrm{NLL}/\partial\log\sigma_n$ falls along the branch.

\paragraph{Factorable set}
For $16\le b\le64$, $g(\theta_b)-b\in[1.7,4.6]$ under both protocols. For
$16\le b\le48$, $\log_2(\sigma_f^2/\sigma_n^2)-b\in[-2.1,-1.1]$ except for
the $24$- and $40$-bit continuation fits ($-0.6$, $+0.3$), and
$\lambda_{\min}(R_{\theta_b})/u$ equals $\sigma_n^2/(u\sigma_f^2)$ within
$0.01$ bits. Over the map's $6174$ trials
$\log_2\lambda_{\max}(R_\theta)\in[3.1,4.9]$, so $h(\theta)$ and $g(\theta)$
are interchangeable, and every failure has $g(\theta)\ge b+3.0$. Of the $488$ points failing the $53$-bit pipeline, $88\%$
fail with entries rounded once and factored at $384$ bits, $96\%$ with
$53$-bit entries factored at $384$ bits, and $419$ fail at pivots
$27$--$29$ of $30$. Where $53$-bit Cholesky succeeds, the NLL error is
about $0.2$ nats for $\lvert x\rvert<1$ and $10^{-3}$ for $7\le x<8$. At
$\theta_{\mathrm{ref}}$, $A$ with float64 entries has
$\lambda_{\min}=-1.1\times10^{-14}$ against $3.9\times10^{-18}$ for exact
entries; $80$-bit entries agree to six digits.

\paragraph{Flat limit}
At $\theta_{\mathrm{ref}}$, $\log_2\kappa_2(K_{t\ell})$ grows by
$13.2$--$14.9$ bits per doubling for $t\ge1$ and by $7$--$10$ for $t<1$,
before the asymptotic regime; $\log_2\kappa_2(A)$ saturates at $72$ bits,
since already at $t=1$, $\lambda_{\min}(K)=2.9\times10^{-18}$ against
$\sigma_n^2=10^{-18}$.

\paragraph{Branches and prediction}
At $r_b=12$, the level of the $53$-bit fits, the fixed-noise branch loses
$11.9$ nats and the free branch $3.77$. The fits from $48$ to $64$ bits are
displaced $0.4$--$0.8$ times as far as the fixed-noise model predicts, and
it over-predicts their loss by a factor $1.8$--$2.9$ (optimizer failures
excluded); the fits from $16$ to $32$ bits are displaced $0.15$--$0.36$
times as far, and the $40$-bit fit roughly opposite to the prediction. The $g(\theta)\le b+\beta_g$ model gives losses
within $4\%$ of the $h(\theta)$ model at every offset except $\beta_{95}$ ($8\%$ at
$60$ bits, $21\%$ at $64$).

\begin{table}[tbhp]
\centering
\caption{Constrained branches of \cref{fig:branch}: fixed-noise loss over
$r_b^2/(2q)$, multiplier of the $\sigma_n$ bound on the fixed-noise branch,
and the free branch's $\sigma_n$ over its floor and loss over the
fixed-noise loss.}
\label{tab:branch}
\footnotesize
\setlength{\tabcolsep}{3pt}
\begin{tabular}{l rrrrrrrrr}
\toprule
$r_b$ & $0.01$ & $0.1$ & $0.3$ & $0.6$ & $1$ & $2$ & $4$ & $12$ & $20$ \\
\midrule
fixed loss $/\,(r_b^2/2q)$ & $0.999$ & $0.991$ & $0.975$ & $0.951$ & $0.923$ & $0.867$ & $0.826$ & $0.889$ & $0.643$ \\
$\sigma_n$-bound multiplier & $0.127$ & $0.103$ & $0.054$ & $-0.011$ & $-0.082$ & $-0.195$ & $-0.239$ & $-0.017$ & $-0.0003$ \\
free $\sigma_n$ / floor & $1$ & $1$ & $1$ & $1.04$ & $1.33$ & $2.06$ & $4.0$ & $47$ & $579$ \\
free loss / fixed loss & $1$ & $1$ & $1$ & $0.994$ & $0.83$ & $0.57$ & $0.41$ & $0.32$ & $0.44$ \\
\bottomrule
\end{tabular}
\end{table}

\paragraph{Optimizers}
Given a zero gradient at a failure, the L-BFGS-B line search returns to the
previous iterate, sees no decrease and stops. In the control two of the
ten starts fail Cholesky at the starting point at $53$ bits, where the
three gradient methods return nothing, and at $72$ bits the Newton method is
held back by a weak Armijo test on a noisy objective. The objective error
of \cref{tab:resolution} follows $c_f\,\kappa_22^{-b}$ with $c_f\approx0.0047$,
almost four orders of magnitude below the constant
$\tfrac12(n+y^\top A^{-1}y)\approx30$ of~\cref{eq:objerr} with $c(n)=1$; at
$96$ bits the continuation chain reaches $1.3\times10^{-6}$, so the
independent fit's excess is optimizer noise.

\begin{table}[tbhp]
\centering
\caption{Objective error at $\theta_{\mathrm{ref}}$, the bound~\cref{eq:resolution}
and the observed independent-protocol displacement along $v_1$, and the
distance $\Delta$ of the Newton-polished fits (``polished'') from the
reference.}
\label{tab:resolution}
\scriptsize
\setlength{\tabcolsep}{2.5pt}
\begin{tabular}{l llllllll}
\toprule
$b$ & $64$ & $72$ & $80$ & $96$ & $128$ & $160$ & $200$ & $256$ \\
\midrule
$\lvert\varepsilon_f\rvert$ & $0.28$ & $6.3\times10^{-4}$ & $1.0\times10^{-5}$ & $4.1\times10^{-11}$ & $5.0\times10^{-21}$ & & & \\
bound & $0.65$ & $0.031$ & $4.0\times10^{-3}$ & $7.9\times10^{-6}$ & & & & \\
observed & $0.32$ & $0.019$ & $2.1\times10^{-4}$ & $3.1\times10^{-3}$ & & & & \\
polished & & & & & $3.7\times10^{-20}$ & $8.8\times10^{-30}$ & $1.2\times10^{-41}$ & $1.5\times10^{-58}$ \\
\bottomrule
\end{tabular}
\end{table}

\paragraph{Generality benchmark}
Every reference passes the $512$-bit check and is the lowest of its
polished modes. Continuation thresholds are within $10$ bits of the
independent ones for the $10^{-2}$ parameter and $10^{-3}$-nat loss
criteria and within $32$ bits at tighter tolerances, the seed-$3$ SE Sobol
case excepted in both. The $53$-bit SE clustered fits lose $6$--$17$ nats; on
the Mat\'ern $5/2$ clustered designs the objective and gradient errors at
$\theta_{\mathrm{ref}}$ are $4\times10^{-4}$--$3\times10^{-2}$ nats and
$3\times10^{-3}$--$0.13$. Of the $584$ benchmark factorizations of
$A_{\theta_{\mathrm{ref}}}$, the $495$ with $\lambda_{\min}(R)\ge2.3u$ all
succeeded,
and $\log_2\lambda_{\max}(R)\in[3.2,5.2]$ ($1.1$ for $d=4$ Sobol). R1 offsets for the
other criteria are $5.4$ (loss $\le10^{-3}$), $3.8$ and $6.9$ (posterior
mean and standard deviation $\le10^{-3}$), $0.6$ (Cholesky success at
$\theta_{\mathrm{ref}}$) and $6.2$, $19.5$, $26.2$
($\lvert\varepsilon_f\rvert\le10^{-3},10^{-6},10^{-9}$). The Mat\'ern $3/2$
counterexample is predicted at $23$ bits. On SE Sobol seed $3$ the
$72$- and $80$-bit fits stop $43$ nats above the reference and polish into
a mode $39$ nats worse, one of the reference multistart's modes; the
continuation protocol passes from $40$ bits. On $d=4$ Sobol the reference
has two length scales on their upper bound and eight modes within $3.1$
nats, and every fit ends in a mode $1.34$ nats above it, except the
$80$-bit fit, which ends in another mode $0.71$ nats above it.
Lowering the floor to $10^{-14}$ moves $\sigma_f$ by $11\%$ and $\ell_2$ by
$5\%$.

\begin{table}[tbhp]
\centering
\caption{Stage B: $\log_2\kappa_2$ of $A$ and $K$ at
$\theta_{\mathrm{ref}}$ and thresholds $b^\ast$ for
$\lVert\delta\theta_{\mathrm{free}}\rVert_\infty\le10^{-2}$ and for a
posterior-mean change $\le10^{-3}$ (m), for the seed-$3$ SE problem
(anchor) with one ingredient changed (radii: cluster radii scaled); $d$ rows: synthetic objective,
clustered (c) or Sobol (s) design.}
\label{tab:stageb}
\footnotesize
\setlength{\tabcolsep}{3pt}
\begin{tabular}{l rrrr @{\hspace{12pt}} l rrrr}
\toprule
case & $A$ & $K$ & $b^\ast$ & m & case & $A$ & $K$ & $b^\ast$ & m \\
\midrule
anchor & $69.6$ & $70.0$ & $74$ & $72$ & $n=50$ & $73.6$ & $101.8$ & $74$ & $72$ \\
radii $\times0.1$ & $69.1$ & $87.2$ & $76$ & $74$ & $d=1$, c & $72.9$ & $237.6$ & $72$ & $53$ \\
radii $\times10$ & $48.3$ & $48.3$ & $53$ & $50$ & $d=1$, s & $72.6$ & $146.3$ & $72$ & $32$ \\
floor $10^{-14}$ & $70.5$ & $70.5$ & $76$ & $76$ & $d=2$, c & $57.4$ & $57.4$ & $61$ & $61$ \\
floor $10^{-6}$ & $49.1$ & $65.5$ & $56$ & $56$ & $d=2$, s & $18.7$ & $18.7$ & $32$ & $32$ \\
obs.\ noise $10^{-6}$ & $49.0$ & $65.3$ & $55$ & $53$ & $d=4$, c & $37.8$ & $37.8$ & $44$ & $44$ \\
$n=20$ & $40.5$ & $40.5$ & $44$ & $44$ & $d=4$, s & $3.9$ & $3.9$ & $>256$ & $>256$ \\
\bottomrule
\end{tabular}
\end{table}

\paragraph{Posterior and active learning}
In the window $\pm0.06$ about the minimizer the $53$-bit fit changes the
posterior mean by $7.3\times10^{-5}$ and the standard deviation by $750\%$
(relative $L^2$; $107\%$ over the domain, average ratio $2.6$); at $64$
bits the average ratios are $1.12$ (domain) and $1.34$ (window), and from
$72$ bits within $2\%$. On the quadratic the reference ends at the cap
$n\sigma_f^2/\sigma_n^2\approx2^{74.5}$ and the $b$-bit fits at
$\log_2(\sigma_f^2/\sigma_n^2)$ between $b-2.5$ and $b-2.0$. The $b$-bit model ($X_b$,
$\theta_b$) evaluated at $200$ bits has final errors $2.8\times10^{-7}$,
$7.5\times10^{-9}$, $9.6\times10^{-7}$, $3.2\times10^{-7}$ and
$1.3\times10^{-5}$ in the columns of \cref{tab:bo-decomp}.

\paragraph{Remedies}
The reference arm's inactive loss is the closed-loop protocol's stopping
floor on two seeds and, on seed $0$, a stop in a second mode ($5.8$ nats;
the mode itself lies $4.2$ nats above the optimum). The $80$-bit
loss falls to at most $10^{-7}$ after a Newton polish. The $64$-bit arm has
$729$ failed evaluations and $6748$ negative predictive variances; jitter
arms and arms from $80$ bits have none. The refit costs $34$--$376$ s; as
fitted it stops $2.1$ and $3.7$ nats short on two inactive designs, which
the polish removes, and on Branin seed $1$, $X_{53}$, it lands $294$ nats
from the optimum. On inactive seed $0$, eleven of fifteen $384$-bit fits
reach the same stationary point (projected gradient below $10^{-52}$), with
recovery error $1.07\times10^{-5}$ against $1.2\times10^{-6}$ for a mode
$4.2$ nats above it. scikit-learn's default has recovery error $0.64$ on
the $n=30$ problem; matched with $\alpha=0$ it loses $23$ nats.

\section*{Acknowledgments}
The author used the large language models Claude (Anthropic) and ChatGPT
(OpenAI) to assist with manuscript preparation and code implementation.
The author assumes responsibility for all content.

\bibliographystyle{siamplain}
\bibliography{references}

\end{document}